\documentclass[conference]{IEEEtran}
\IEEEoverridecommandlockouts
\renewcommand{\baselinestretch}{0.99}
\usepackage{cite}
\usepackage{amsmath,amssymb,amsfonts}
\usepackage{graphicx}
\usepackage{textcomp}
\usepackage{algpseudocode}
\usepackage{algorithm}
\usepackage{multirow}
\usepackage{xcolor}
\usepackage{comment}
\usepackage{url}
\usepackage{hyperref}
\usepackage{tabularx}
\usepackage{booktabs}
\usepackage[most]{tcolorbox}
\def\BibTeX{{\rm B\kern-.05em{\sc i\kern-.025em b}\kern-.08em
    T\kern-.1667em\lower.7ex\hbox{E}\kern-.125emX}}
\begin{document}



\title{Comparative Analysis Based on DeepSeek, ChatGPT, and Google Gemini: Features, Techniques, Performance, Future Prospects}

\author{\IEEEauthorblockN{Anichur Rahman\IEEEauthorrefmark{1}, Shahariar Hossain Mahir\IEEEauthorrefmark{2}, Md Tanjum An Tashrif\IEEEauthorrefmark{3}, Airin Afroj Aishi\IEEEauthorrefmark{4}, Md Ahsan Karim\IEEEauthorrefmark{5}, \\Dipanjali Kundu\IEEEauthorrefmark{6}, Tanoy Debnath\IEEEauthorrefmark{7},  Md. Abul Ala Moududi\IEEEauthorrefmark{8}, 
and MD. Zunead Abedin Eidmum\IEEEauthorrefmark{9}}

\IEEEauthorblockA{
\textit{Deptment of Computer Science and Engineering, National Institute of Textile Engineering and Research (NITER)},\\
\textit{Constituent Institute of Dhaka University, Savar, Dhaka-1350}\\
\textit{Department of Computing and Information System, Daffodil International University, Savar, Dhaka, Bangladesh},\\
\textit{Department of Computer Science, Stony Brook University, Stony Brook, NY, USA},\\
\textit{Department of Internet of Things and Robotics Engineering, Bangabandhu Sheikh Mujibur Rahman Digital University, Bangladesh},\\
anis\_cse@niter.edu.bd\IEEEauthorrefmark{1},
mahihossain114@gmail.com\IEEEauthorrefmark{2},
mtashrif20@niter.edu.bd\IEEEauthorrefmark{3},
airinafroj93@gmail.com\IEEEauthorrefmark{4},
makarim11@niter.edu.bd\IEEEauthorrefmark{5},\\
dkundu@niter.edu.bd\IEEEauthorrefmark{6},
tadebnath@cs.stonybrook.edu\IEEEauthorrefmark{7},
mmoududi21@niter.edu.bd\IEEEauthorrefmark{8},
1801031@iot.bdu.ac.bd\IEEEauthorrefmark{9}}}

\maketitle

\begin{abstract}
Nowadays, DeepSeek, ChatGPT, and Google Gemini are the most trending and exciting Large Language Model (LLM) technologies for reasoning, multimodal capabilities, and general linguistic performance worldwide. DeepSeek employs a Mixture-of-Experts (MoE) approach, activating only the parameters most relevant to the task at hand, which makes it especially effective for domain-specific work. On the other hand, ChatGPT relies on a dense transformer model enhanced through reinforcement learning from human feedback (RLHF), and then Google Gemini actually uses a multimodal transformer architecture that integrates text, code, and images into a single framework. However, by using those technologies, people can be able to mine their desired text, code, images, etc, in a cost-effective and domain-specific inference. People may choose those techniques based on the best performance. In this regard, we offer a comparative study based on the DeepSeek, ChatGPT, and Gemini techniques in this research. Initially, we focus on their methods and materials, appropriately including the data selection criteria. Then, we present state-of-the-art features of DeepSeek, ChatGPT, and Gemini based on their applications. Most importantly, we show the technological comparison among them and also cover the dataset analysis for various applications. Finally, we address extensive research areas and future potential guidance regarding LLM-based AI research for the community.
\end{abstract}

\vspace{2mm}

\section{Introduction}
Large Language Models (LLMs) have completely reshaped the fields of natural language processing and artificial intelligence. These models now allow computers not only to process and generate human language but also to engage in reasoning based on it \cite{wu2023next}. In the past few years, tools like DeepSeek, ChatGPT, and Gemini have further advanced these capabilities. By integrating specialized domain knowledge, using reinforcement learning from human feedback (RLHF), and supporting multimodal inputs, they have broadened the scope of AI applications \cite{yao2024survey, rahman2023federated}. Today, such technologies are making their mark in areas like healthcare, finance, education, and customer service, where they help deliver personalized responses and tackle complex analytical tasks.

\subsection{Motivations}
The rapid evolution of large models is in large part a response to increasing demand for AI systems to manage realistic, complex challenges to high precision \cite{jurenka2024towards}. DeepSeek is one such that embraces a mixture-of-experts (MoE) approach, in that it accurately selects only the most relevant parameters to be used during prediction. This targeted activation reduces computational expenses and boosts efficiency, especially for applications focused on specific domains \cite{cai2024survey}. Meanwhile, ChatGPT enhances its conversational skills by incorporating reinforcement learning from human feedback (RLHF), allowing it to generate responses that are both context-aware and fluent \cite{mondillo2024basal}. In contrast, Gemini sets itself apart with a multimodal design that merges text, images, and audio, enabling it to process and produce outputs across different data types. These breakthroughs not only improve overall model performance but also pave the way for innovative applications of LLMs in specialized areas \cite{rahman2022sdn}.

\subsection{Related Study}
Recent studies have shed considerable light on how modern large language models (LLMs) are developed and performed. For example, Smith et al. \cite{smith2024deepseek} offer a detailed look at DeepSeek’s architecture, showing that its mixture-of-experts (MoE) method boosts both efficiency and performance when handling queries in specific domains. Similarly, Rojas and Kim \cite{rojas2025chatgpt} explore ChatGPT’s ability to work across multiple languages and sustain coherent, context-aware conversations on various topics, achieving high accuracy in its responses.

Johnson et al. \cite{johnson2025gemini} focus on the multimodal transformer design of Gemini, explaining how its blend of text, code, and visual data supports advanced reasoning across different types of inputs in complex tasks. In a related study, Park and Gupta (2024) examine the cross-modal alignment techniques used in Gemini, which enhance the model’s capacity to merge diverse information sources effectively. Meanwhile, Anderson et al. \cite{anderson2024chatgpt} look into scaling strategies and the use of reinforcement learning from human feedback (RLHF) in ChatGPT, noting clear improvements in dialogue generation and overall user interaction.
Together, these works offer a rich, multi-layered perspective on current methods and benchmarks in state-of-the-art LLMs.

In addition, Chang et al. \cite{chang2024survey} provide a broad overview of several leading LLMs—including GPT, LLaMA, and PaLM—discussing their architectures, key contributions, limitations, the datasets they use, and how they perform on standard benchmarks. Qin et al. \cite{qin2024large} also summarizes the architectures and contributions of influential models such as BERT, GPT-3, and LLaMA, contrasting their strengths, weaknesses, and performance across a broad array of natural language processing tasks. Lastly, Sindhu et al. \cite{sindhu2024evolution} point out present applications, challenges, and directions for future research in LLMs, with a focus on continued transformer architecture innovation and performance metric refinement to advance further.
A selection of papers has been shown in Table \ref{tab:lit_review}, to compare previous studies. The main contribution of the paper is:

\begin{itemize}
    \item  We develop a comprehensive comparative analysis of DeepSeek, ChatGPT, and Gemini, focusing on their architectures, training methodologies, and domain-specific applications with proper explanation.\vspace{1.5mm}

    \item The authors focus on the methods and materials for collecting and searching appropriate data based on DeepSeek, ChatGPT, and Gemini with technical comparison among them. \vspace{1.5mm} 
    
    \item Further, we implement a performance benchmarking framework that evaluates these models across standardized metrics such as accuracy and reasoning capability.\vspace{1.5mm}

     \item Finally, we implement a detailed discussion on the limitations and future opportunities for LLMs, emphasizing improvements in explainability and multimodal integration.
\end{itemize}\vspace{2mm}

\textbf{Organization of the Survey:} 
The remainder of this study is organized as follows. Section II describes the Materials and Methods employed by DeepSeek, ChatGPT, and Gemini. Section III details these models' state-of-the-art features and technical innovations, while Section IV compares their performance, training datasets, and evaluation metrics. Section V presents our Popular Datasets for Developing the DeepSeek, ChatGPT, and Gemini Applications. Further, Section VI discusses performance analysis. Challenges and future direction have been covered by Section VII. Additionally, Sections VIII \& IX explain the limitations and discussion. Finally, we conclude the paper in Section X.


\begin{table*}[ht]
\centering
\caption{Comparison of Recent Studies on ChatGPT, Gemini, and DeepSeek}
\label{tab:lit_review}
\begin{tabular}{p{2cm} p{3cm} p{3cm} p{3cm} p{2.5cm} p{2.5cm}}
\hline
\textbf{Study} & \textbf{Models Compared} & \textbf{Methodology/Evaluation} & \textbf{Key Findings} & \textbf{Strengths} & \textbf{Weaknesses} \\
\hline
Ahmed et al.\cite{ahmed2024gemini} & Gemini vs. GPT-4V & Mixed qualitative/quantitative analysis on visual reasoning tasks & Gemini offers concise visual reasoning, though simple details (e.g., clock times) may be misrepresented & Effective multi-modal integration & Occasional systematic errors \\
\hline
Liu et al.\cite{liu2024deepseek} & DeepSeek-V3 vs. Llama 3.1, GPT-4o, Claude 3.5 & Benchmark tests on reasoning tasks and cost analysis & Achieves competitive reasoning at a fraction of the cost with extended context capabilities & High cost efficiency; long context window & Inference speed drops under heavy demand \\
\hline
Shao et al.\cite{shao2024deepseekmath} & DeepSeekMath vs. baseline models (e.g., ChatGPT) & Reinforcement learning with chain-of-thought verification on mathematical tasks & Significant improvements on complex math problems through enhanced reasoning & Robust math reasoning and step-by-step internal checks & Limited generalizability beyond mathematics \\
\hline
Wang et al.\cite{wang2024math} & Multiple LLMs (including ChatGPT variants, DeepSeek) & Use of reward models for step-by-step verification on math and coding tasks & Reinforcement learning improves accuracy and transparency in reasoning & Better error correction; clear reasoning traces & Additional computational overhead and latency \\
\hline
Liu et al.\cite{liu2024deepseek} & DeepSeek-V2 vs. open-source/commercial models & Evaluation via a mixture-of-experts architecture and token benchmarks & Strong performance with excellent resource usage and scalability & Economical and efficient design & Challenges in expert balancing \\
\hline
Cheng et al.\cite{cheng2024vision} & Vision-Language models (Gemini) vs. ChatGPT baseline & Reflective chain-of-thought prompting on multi-modal benchmarks & Reflective methods enhance visual and textual reasoning & Improved integration of visual and textual data & Increased processing times and higher computational cost \\
\hline
Xu et al.\cite{xu2024survey} & ChatGPT, Gemini, DeepSeek & Efficiency and energy consumption benchmarks & Gemini shows slightly higher energy efficiency; DeepSeek excels in cost, while ChatGPT remains robust & Clear trade-offs in energy and cost & Variability due to dynamic energy benchmarks \\
\hline
Sun et al.\cite{sun2024openomni} & ChatGPT, Gemini & Hybrid evaluation combining open-source and proprietary strategies for conversation & Hybrid approaches can leverage strengths from each model; open-source components boost transparency & Innovative integration; improved robustness & Increased system complexity \\
\hline
Peng et al.\cite{peng2024securing} & ChatGPT, Gemini & Controlled study on safety and bias using a set of standardized prompts & ChatGPT and Gemini better mitigate bias than DeepSeek, though with less flexibility & In-depth safety and ethical evaluation & Limited sample size and prompt-specific findings \\
\hline
\end{tabular}
\end{table*}

\section{Materials and Methods}
\begin{figure}[H]
    \centering    \includegraphics[width=0.46\textwidth]{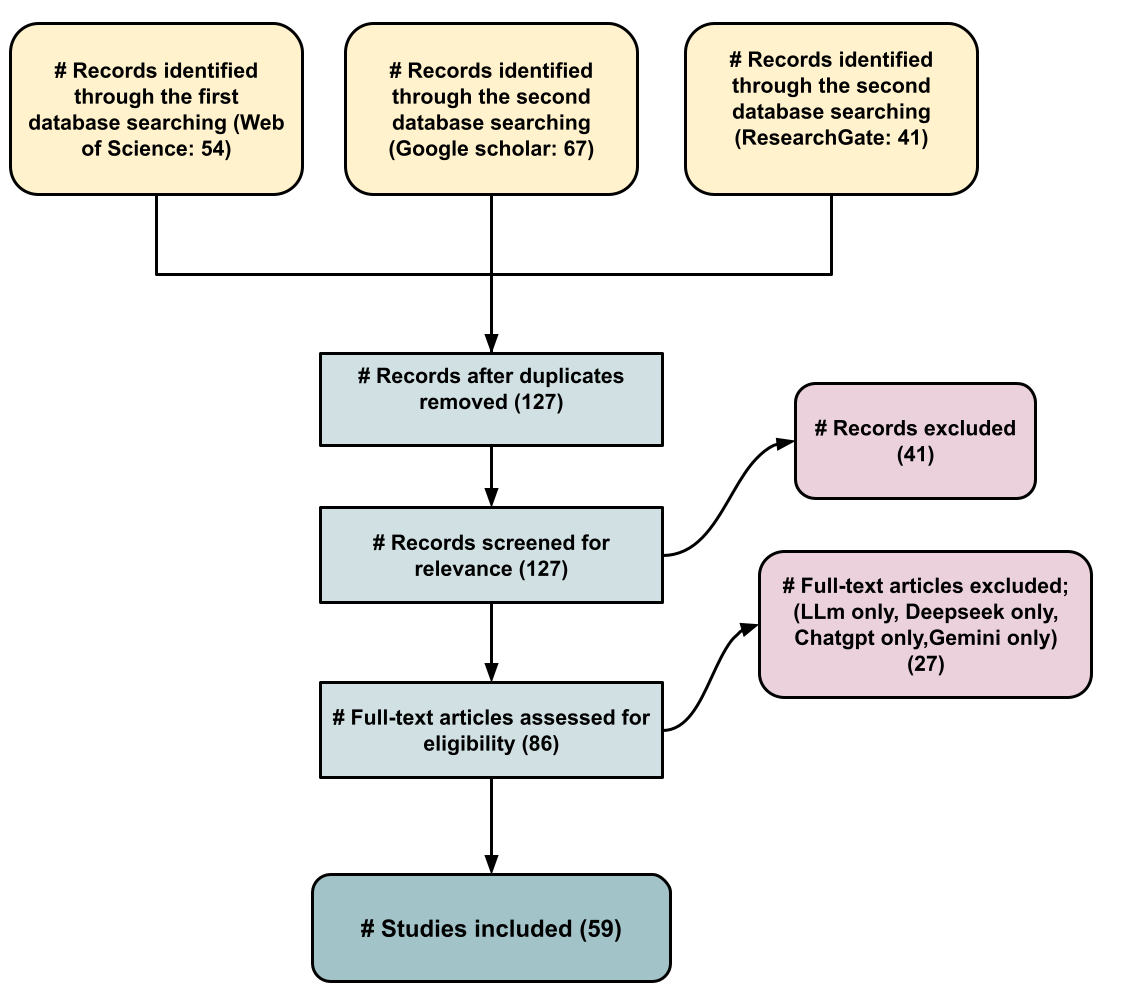} 
    \caption{Flow diagram of the paper selection procedure according to PRISMA}
    \label{fig:prisma}
\end{figure}

The research strategy for this review paper was designed to comprehensively evaluate the strengths, limitations, and comparative capability of DeepSeek, ChatGPT, and Gemini. Transparency and reproducibility were ensured in the research process by adopting the PRISMA guideline\cite{valle2024artificial} in the selection process of the papers. Web of Science, Google Scholar, and ResearchGate from the PRISMA diagram (Figure \ref{fig:prisma}) were used to search for the related studies. After removing duplicates, 127 records were screened for relevance, with an additional rigorous evaluation of 86 full-text articles. Of these, 59 papers reached the final analysis based on some inclusion and exclusion criteria \cite{hasan2021normalized}.\vspace{2mm}

Papers were chosen based on their presentation of architectural specifications of AI models, cross-domain validity experiments, or comparison by relative parity with current systems. Papers omitted were those that demonstrated a lack of methodological soundness, were too narrow in their focus on singular models without comparative outlook, or failed tests of applicability in the real world to domains of logical reasoning, code generation, or multimodal integration. This rigorous curation enabled the presentation of valid, actionable research and the conduct of firm, contextually driven analysis.

\subsection{Research Question}

The overall research question of the project is analyzing the strengths, weaknesses, and trade-offs among DeepSeek, ChatGPT, and Gemini across various domains and tasks. In particular, the project investigates how the three models compare in terms of accuracy, logical and numerical reasoning, programming ability, and response generation. The comparison also examines how the underlying architectures of the models—DeepSeek's Mixture-of-Experts (MoE)\cite{neha2025survey}, ChatGPT's dense transformer model with reinforcement learning from human feedback (RLHF)\cite{wang2024rlhf}, and Gemini's multimodal transformer—affect their usability, efficiency, and performance in real tasks\cite{team2024gemini, islam2021blockchain}.

\subsection{Paper Collection}

The paper collection was conducted in a systematic way with the PRISMA protocol openly. The initial searches produced 54 records via Web of Science, 67 records via Google Scholar, and 41 records via ResearchGate. After removing duplicates, 127 papers were screened against relevance criteria, including whether or not they had provided comparative insight into model performance, architecture, or domain-specific applications.

We excluded research that did not provide thorough information on model assessments, were solely on one model with no comparison, or only considered subgroups of models (e.g., only assessing Gemini's multimodal functionality with no comparison to DeepSeek or ChatGPT). Following relevance screening, 86 full-text papers were screened for eligibility based on more specific criteria, including the presence of performance evaluation metrics, explanation of domain-specific tasks, and discussion of real-world application issues. Finally, 59 articles fulfilled all the criteria and were selected for subsequent analysis. Figure 1 displays this process via the PRISMA diagram for transparency and reproducibility \cite{rahman2023icn}.

\subsection{Data Sets and Data Sources}
Datasets for training and testing of the models differed, and general-purpose and domain-specific sources were used. Primary sources for general linguistic coverage included Common Crawl and WebText, and BooksCorpus was used for long-form structured text data\cite{sachdeva2024train, islam2021sgbba}. Domain-specific corpora like PubMed and arXiv were essential for testing the generation of scientific and medical text. GitHub and Stack Overflow datasets enabled code-based tasks. Gemini's testing also needed multimodal datasets, like LAION-400M, which consist of image-text pairs testing the model's capability in cross-modal content generation and processing\cite{bi2024deepseek}. Private datasets were also used for testing domain-specific use cases in domains like finance and law. This variation in the dataset creation allowed comprehensive testing of each model's capabilities on heterogeneous tasks.

\subsection{Dataset Paper Collection}

The dataset-gathering process was framed in such a manner that each model was tested based on its individual design and specialization. DeepSeek, being a specialized model, relied on curated medical, legal, and financial corpora. ChatGPT's dataset was more general-purpose, with general web text, curated dialogue transcripts, and open-source code repositories. Gemini was founded on a multimodal dataset that merged text, code, and visual data, enabling robust cross-modal reasoning and content generation. This categorization ensured that each model’s unique strengths were considered during the evaluation process, allowing for meaningful performance comparisons across different contexts.

\subsection{Different Performance Testing}
\begin{figure*}[!htbp]
\centerline{\includegraphics[scale=0.30]{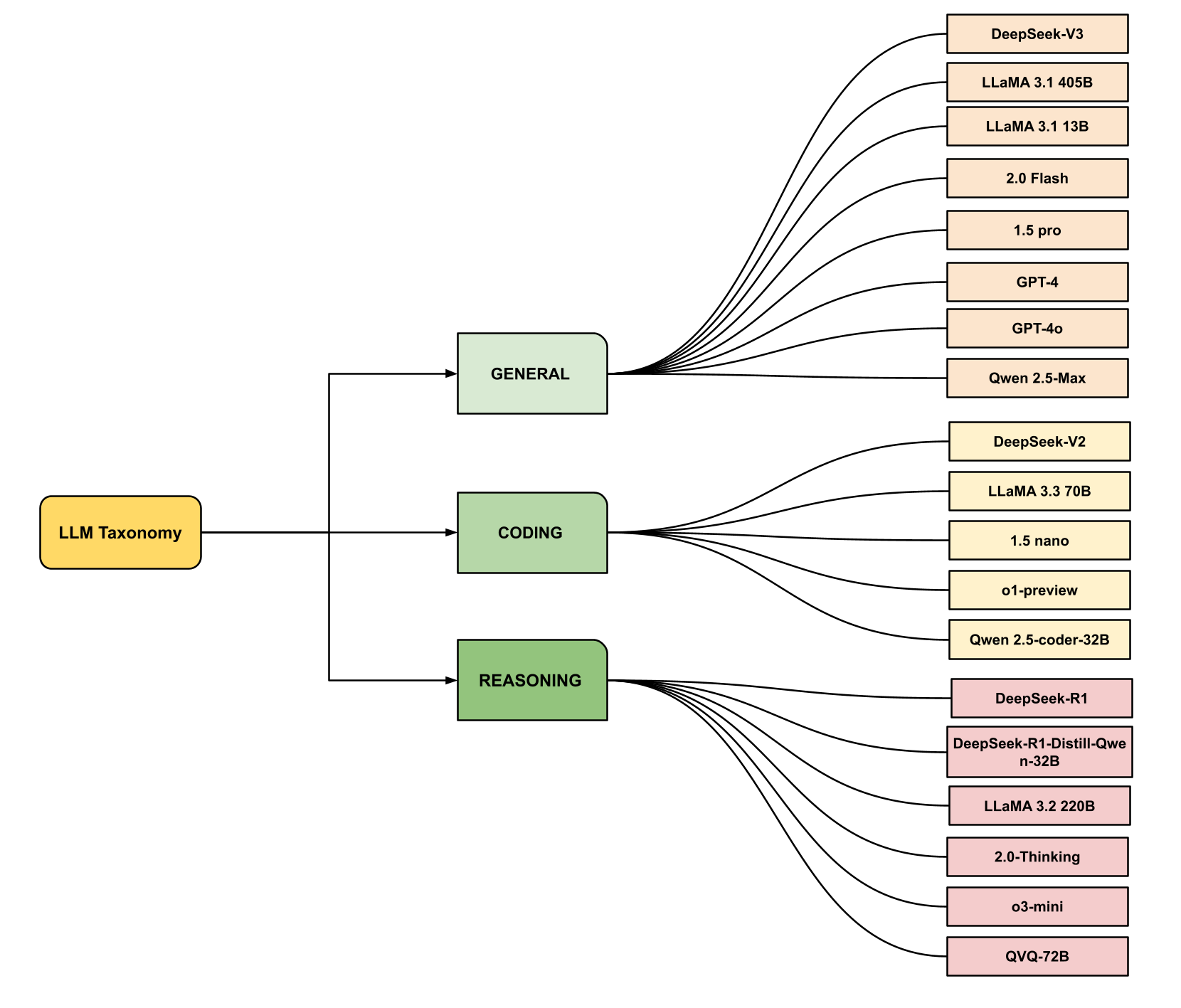}}
\caption{LLM Taxonomy of Tested Models}
\label{fig:texo}
\end{figure*}

Various performance areas were explored to provide a detailed evaluation of the weak points and strong areas of each model\cite{coignion2024performance}, \cite{zhang2024o1}. Quantitative and logical reasoning tests, which were evaluated against benchmarks like MMLU, put the mathematical and analytical problem-solving skills of the models to the test. Reasoning tests involving complex and multi-step complexities were aimed at evaluating the models' capacity to answer context-dependent, multi-step questions. Coding proficiency was evaluated on HumanEval benchmarks that measured the models' ability to generate working code, debug, and generate detailed explanations\cite{hanci2024assessment}. Usability testing for everyday use focused on conversational flow, contextual recall, and relevance of answers in real-world simulation tests. Further, the ability of the models to function with multilingual prompts was evaluated to assess their suitability for different languages and cultural contexts. The taxonomy of the paper is shown in Figure \ref{fig:texo}.

\subsection{Model Architectures and Techniques:}
The architecture of each model was scrutinized to determine how design decisions influenced performance. DeepSeek's Mixture-of-Experts (MoE) architecture conditionally activated pertinent parameters at inference, modulating computational cost to domain-specific prompts. ChatGPT's dense transformer model, having been optimized with RLHF, allowed it to preserve contextual coherence and conversational fluency, especially in dynamic dialogue environments. Gemini's multimodal transformer model design facilitated easy integration of text, code, and image data with robust support for cross-modal reasoning and content creation. By comparing these architectures, the work highlighted each model's unique strengths and limitations across different application scenarios.

\subsection{Computational Resources:}

All experiments were conducted in a consistent, high-performance computing environment to ensure fairness and reproducibility. High-performance GPUs were used to provide reliable computational resources to all models. Standardized setups—fixed input token lengths, batch sizes, and memory allocs—were used to remove possible biases in performance comparisons. Scalability tests were further done by changing computational loads to test the performance stability of the models under actual usage scenarios. This controlled setup guaranteed that the results of testing were reliable and a true representation of the models' actual capabilities.

\section{DeepSeek, ChatGPT, and Gemini  Applications: State-of-the-art Features}

\subsection{Early Features before Releasing Generative Artificial Intelligence (AI) Tools}

Prior to the advent of contemporary generative AI, research in language models was driven by advances that laid the groundwork for today's sophisticated systems. The initial methods relied on static word representations and rule-based processing, as in the case of Word2Vec and GloVe models, which brought in dense vector embeddings to represent semantic relationships\cite{ertel2024introduction, rahman2025impacts}. These methods, while handicapped by the lack of context-dependent meaning, were important contributions to research in compact representation of language and scalability. This initial research highlighted the promise of exploiting large datasets for tasks in natural language, a concept that would come to underpin the domain-specific optimizations of models such as DeepSeek\cite{zhang2021study}. This initial period, characterized by the discovery of successful embedding strategies and early transformer concepts, laid the foundation for the evolution of more active, generative architectures\cite{cao2022new}.

\subsection{After Releasing Modern Generative Artificial Intelligence (AI) Tools}
The introduction of modern generative AI tools marked a paradigm shift in natural language processing by enabling models to generate coherent, contextually rich outputs\cite{gao2024artificial}. In this era, DeepSeek, ChatGPT, and Google Gemini each demonstrate state-of-the-art features tailored to their unique design philosophies.

\subsubsection{Features of DeepSeek Tools}

DeepSeek tools, in particular the R1 version take advantage of a mixture-of-experts architecture that dynamically activates model parameters relevant to the input at hand. On top of optimizing compute efficiency, such an architecture provides increased robustness for domain-specific applications—i.e., medical, legal, and financial—by sustaining performance under computational-resource limitations. DeepSeek R1's specialized design means that it will perform exceptionally well at specialized tasks without incurring high operational expenditure\cite{guo2025deepseek, bhuiyan2023iot}.

\subsubsection{Features of ChatGPT Tools}

ChatGPT, on the other hand, builds upon a dense transformer framework enriched by reinforcement learning from human feedback (RLHF). This combination enables rapid response times and agile conversational interactions, making ChatGPT highly effective in real-time dialogue scenarios\cite{zhang2024exploring}. Its architecture is particularly resilient in preserving context over extended conversations, even though it occasionally encounters challenges with complex or ambiguous prompts. The strength of ChatGPT lies in its ability to deliver fast, coherent, and adaptable text generation for a wide array of general-purpose applications\cite{zhang2024latest}.

\subsubsection{Features of Gemeni Tools}

Google Gemini marks a notable advance in multimodal integration. Its transformer-based architecture is designed to process text, code, and visual data simultaneously, thereby broadening its applicability to tasks such as generative content creation and cross-modal analysis\cite{imran2024google}. Despite requiring additional computational resources, Gemini consistently demonstrates robust performance across a variety of data types, underscoring its capability to manage sophisticated, combined tasks. This multimodal functionality makes it particularly well suited for applications that rely on the integration of diverse information sources\cite{rane2024gemini}.

\subsection{Comparison among DeepSeek, ChatGPT, and Gemeni based on their Features}

As shown in Table \ref{tab:comparison}, the side-by-side comparison of these models highlights distinct strengths. DeepSeek R1 is distinguished by its high efficiency and domain-specific robustness, which are achieved through its mixture-of-experts architecture. In contrast, ChatGPT is celebrated for its rapid response times and conversational agility, attributes that stem from its dense transformer model enhanced with reinforcement learning from human feedback. Google Gemini, meanwhile, excels in supporting multiple modalities—text, code, and images—delivering strong performance in cross-domain scenarios even with its increased computational demands.

\begin{table*}[h]
\centering
\caption{Comparative Analysis of DeepSeek R1, ChatGPT, and Google Gemini.}
\begin{tabular}{|p{3cm}|p{4cm}|p{4cm}|p{4cm}|}
\hline
\textbf{Feature} & \textbf{DeepSeek R1} & \textbf{ChatGPT} & \textbf{Google Gemini} \\
\hline
Architectural Design & Mixture-of-experts (MoE) that selectively activates parameters & Dense Transformer enhanced with RLHF & Multimodal Transformer integrating text, code, and images \\
\hline
Efficiency & Optimized for cost-effective, domain-specific processing & Fast response times for real-time interactions & Robust handling of diverse data; higher resource demand \\
\hline
Speed & Delivers consistent performance in specialized tasks & Excels in rapid conversational exchanges & Moderately fast; performance tied to multimodal complexity \\
\hline
Resilience & Maintains stable output under resource constraints in targeted applications & Preserves context over extended dialogues & Demonstrates high resilience across varied data types \\
\hline
Specialization & Tailored for sectors like healthcare, law, and finance & Versatile for general-purpose and conversational tasks & Ideal for integrated, cross-modal analysis and creative applications \\
\hline
\end{tabular}
\label{tab:comparison}
\end{table*}

\section{DeepSeek, ChatGPT, Gemeni: Technical Comparison}

The rapid advancement of large language models (LLMs) has been driven by breakthroughs in training architectures, optimization techniques, and data-driven learning approaches. Understanding the technical foundations of these models requires examining their evolution, from early pre-trained models to the latest state-of-the-art systems. This section explores the key innovations that have shaped DeepSeek, ChatGPT, and Gemini, beginning with an overview of early pre-trained models.

\subsection{Early Pre-trained Models}

\textcolor{black}{The foundation for current large language models (LLMs) was established by a number of early pre-trained models before the emergence of DeepSeek, ChatGPT, and Gemini. These models made significant contributions to large-scale training, self-supervised learning, and transformer architectures \cite{zhao2023survey}. Some of the most significant early models are as follows:}
\subsubsection{\textbf{Word Embedding Models (Pre-2018)}}
{\color{black}Instead of generative text features, the primary focus of early NLP models was on static word representations. One of the most noteworthy innovations was Word2Vec \cite{church2017word2vec}, created by Google in 2013. In order to capture the semantic links between words, Word2Vec presented an innovative method to express words as continuous vector embeddings. It used two important architectures, Skip-gram \cite{mccormick2016word2vec} and Continuous Bag-of-Words (CBOW) \cite{xia2023continuous}, which allowed models to anticipate words in a specific context. For example, if the vector for ``king'' is modified by deleting the vector for ``man'' and adding the vector for ``woman'', the result is a vector representation that is quite similar to ``queen" \cite{mikolov2013efficient}. This shows how Word2Vec identifies relevant semantic connections between words based on their context. However, Word2Vec's primary drawback was that it produced static embeddings, which meant that words always had the same vector representation regardless of the context in which they were used.
\vspace{2mm}

In 2014, Stanford introduced GloVe (Global Vectors for Word Representation), which came after Word2Vec. GloVe used a matrix factorization technique based on global word co-occurrence statistics, in contrast to Word2Vec, which depended on local context windows. As a result, it was able to more accurately represent the connections between words across whole corpora \cite{pennington2014glove}. GloVe's efficacy in tasks demanding sophisticated understanding was limited by its inability to produce context-dependent word representations, regardless of significant advancements.
\vspace{2mm}

 In 2016, Facebook AI Research built FastText \cite{joulin2016fasttext}, which improved word embeddings by integrating subword information. This enables the model to handle unusual words more effectively by dividing them down into smaller components (subwords), resulting in better performance on languages with complicated morphology \cite{bojanowski2017enriching}. The influence of these early word embedding models was significant, demonstrating the relevance of pre-trained word representations and paving the way for more complex NLP techniques.
}
\vspace{2mm}
\subsubsection{\textbf{Contextualized Language Models (2018-2019)}}
{\color{black}
Contextualized language models, which produced dynamic representations depending on sentence context, were developed in response to the shortcomings of static word embeddings. AllenNLP's 2018 introduction of ELMo (Embeddings from Language Models) \cite{reimers2019alternative} was one of the first significant innovations. ELMo created contextualized word embeddings using deep bidirectional LSTMs (Long Short-Term Memory networks) \cite{zhang2015bidirectional}. ELMo produced distinct representations for the same word based on its usage, in contrast to Word2Vec and GloVe, which gave a word the same vector regardless of its meaning. Especially for challenges using polysemous words like ``bank'' (financial institution vs. riverbank) \cite{Peters2018DeepCW}, this was a significant improvement. However, ELMo was less scalable and computationally costly due to its reliance on LSTMs rather than transformers.
\vspace{2mm}

ULMFit (Universal Language Model Fine-tuning for Text Classification), which was released by fast.ai in 2018, was another significant model from this era. ULMFit refined a pre-trained LSTM-based language model on downstream tasks, demonstrating the efficacy of transfer learning in NLP. This had an impact on later transformer-based architectures and greatly increased the effectiveness of language model training \cite{howard2018universal}. These models had a significant impact because they set the stage for the following transformer revolution.}
\vspace{2mm}
\subsubsection{\textbf{Transformer-based Pre-trained Models (2018-2019)}}
{\color{black}
The emergence of transformer-based architectures, which enhanced stability and performance by substituting self-attention mechanisms for LSTMs, was the most important development in NLP. Instead of using unidirectional context prediction, Google's groundbreaking BERT (Bidirectional Encoder Representations from Transformers) model uses bidirectional attention. Because BERT could analyze text in both directions, it was able to get a broader contextual knowledge of words than earlier models. It was trained using Masked Language Modeling (MLM) and Next Sentence Prediction (NSP), which helped it to excel at tasks like question answering and sentiment analysis \cite{devlin2018bert}. However, BERT's application for conversational AI was limited because it was not intended for text production.
\vspace{1mm}

Using a different strategy, OpenAI's GPT-1 (Generative Pre-trained Transformer) was released the same year and concentrated on autoregressive text production. GPT-1 generated text word each word using unidirectional decoding as opposed to BERT's bidirectional attention. The BooksCorpus dataset was used to train the model, enabling it to generate text that is coherent and fluid \cite{radford2018improving}. But with just 117 million parameters, GPT-1 was comparatively small, which limited its capacity for generalization.}
\vspace{2mm}

\subsubsection{\textbf{Scaling Up with Large-Scale Pre-trained Models (2019-2020)}}
{\color{black}
Models expanded notably in size, strength, and multimodality by 2019–2020. With 1.5 billion parameters, OpenAI's 2019 release of GPT-2 represented a significant advancement. In contrast to its predecessor, GPT-2 was capable of zero-shot learning, which allowed it to finish NLP tasks without the need for manual fine-tuning \cite{radford2019language}. However, OpenAI first delayed its distribution because of worries about abuse and false information. Later instruction-tuned models were inspired by Google's T5 (Text-to-Text Transfer Transformer), which was introduced about the same time and viewed every NLP task as a text-to-text problem \cite{raffel2020exploring}.
\vspace{2mm}

The most transformative model of this era was GPT-3, released by OpenAI in 2020, which scaled up to 175 billion parameters. GPT-3 demonstrated few-shot and zero-shot learning capabilities, making it highly versatile for generating text \cite{brown2020language}. It also marked the beginning of commercial AI services, leading to applications like ChatGPT. These advancements directly influenced the development of DeepSeek, ChatGPT, and Gemini, shaping modern generative AI.
}

\begin{table*}[h]
    \centering
    \renewcommand{\arraystretch}{1.3} 
    \caption{Technical Comparison of DeepSeek, ChatGPT, and Gemini}
    \label{tab:comparison}
    \begin{tabular}{|p{3.5cm}|p{3.5cm}|p{3.5cm}|p{3.5cm}|}
        \hline
\textbf{Feature} & \textbf{DeepSeek (DeepSeek AI)} & \textbf{ChatGPT (OpenAI)} & \textbf{Gemini (Google DeepMind)} \\
        \hline
        \textbf{Architecture} & Mixture of Experts (MoE) & Dense Transformer (GPT-4) & Multimodal Transformer \\
        \hline
        \textbf{Training Data} & Chinese-centric data + multilingual support & OpenAI proprietary data + web data & Google-scale datasets (text, images, audio, video) \\
        \hline
        \textbf{Training Efficiency} & Highly efficient due to MoE architecture & High resource requirements & Balanced efficiency for multimodal tasks \\
        \hline
        \textbf{Scalability} & Open-source model, highly customizable & Large-scale deployment via API & Integrated into Google ecosystem (e.g., Bard, Search) \\
        \hline
        \textbf{Computational Efficiency} & More efficient; fewer active parameters & Requires significant computational resources & Optimized for multimodal processing \\
        \hline
        \textbf{Multimodal Capabilities} & Primarily text-based & Limited (GPT-4V introduced vision support) & Fully multimodal (text, images, audio, video) \\
        \hline
        \textbf{Performance} & Excels in logical reasoning and problem-solving & Superior in natural language understanding and generation & Advanced in multimodal data processing and generation \\
        \hline
        \textbf{Availability} & Open-source & Commercial (OpenAI API) & Integrated into Google products and services \\
        \hline
        \textbf{Strengths} & Cost-effective, scalable, and efficient & Versatile and robust for NLP tasks & Comprehensive multimodal integration \\
        \hline
        \textbf{Ideal Use Cases} & Resource-efficient applications, specialized reasoning tasks & Conversational AI, content creation, and general NLP tasks & Multimedia content creation, cross-modal analysis, and multimodal AI tasks \\
        \hline
    \end{tabular}
\end{table*}

\subsection{Generative Pre-trained Models}
{\color{black}The generative pre-trained model family includes contemporary large language models like ChatGPT, DeepSeek, and Gemini. These models are pre-trained on large text datasets and then adjusted for particular applications. This is a two-step process. Every model has unique architectures and capabilities.}
\subsubsection{\textbf{ChatGPT (OpenAI)}}
{\color{black}
From GPT-3, ChatGPT developed into GPT-3.5 and GPT-4. To improve user interactions, it makes use of RLHF (Reinforcement Learning from Human Feedback) and instruction tailoring \cite{ouyang2022training}. ChatGPT is well-known for its strong text-based AI features, which make it ideal for use in conversational AI, customer service, and coding assistance. Its capacity to produce logical, context-aware language across a range of areas is its fundamental strength. However, ChatGPT's absence of multimodal capabilities—which were eventually added in GPT-4 Vision—was a significant drawback, especially in its early iterations.
}
\vspace{2mm}

\subsubsection{\textbf{DeepSeek}}
{\color{black}
A significant open-source substitute for GPT-based models, especially in the Chinese AI research community, is DeepSeek. A Mixture of Experts (MoE) architecture was introduced by the DeepSeek R1 model, which only activates a portion of its parameters for each query. Compared to dense models like ChatGPT, DeepSeek is hence more computationally efficient. Its cost-effective scaling, which enables effective deployment in a variety of tasks, is its main advantage. However, it has less generalization across languages, as it is primarily optimized for certain domains \cite{mercer2025brief}.
}
\vspace{2mm}

\subsubsection{\textbf{Gemini (Google DeepMind)}}
{\color{black}
PaLM 2 was replaced in 2023 by Gemini, a product of Google DeepMind. Gemini is a completely multimodal model, which means it can process and produce text, images, audio, and video, in contrast to ChatGPT and DeepSeek. This makes it perfect for uses like scientific research and AI-powered content production that call for cross-modal understanding. In comparison to ChatGPT, Gemini still has optimization issues in applications that are solely text-based, despite its notable advantages in multimodal learning \cite{team2023gemini}.
}

\subsection{Techniques Comparison: DeepSeek vs. ChatGPT vs. Gemini}
{\color{black}
The distinctive designs of architecture, training efficiencies, and performance capabilities of DeepSeek, ChatGPT, and Gemini—each suited to particular applications and use cases—are revealed by a comparative study of their methods. Their main characteristics are briefly outlined in Table \ref{tab:comparison}, and their particular advantages and disadvantages are covered in more detail in the discussion that follows.

\subsubsection{\textbf{Architecture}}
{\color{black}
\begin{itemize}
    \item \textbf{DeepSeek:} DeepSeek utilizes a Mixture-of-Experts (MoE) architecture that mainly activates a portion of its parameters during inference. This architecture increases computational efficiency while significantly reducing training costs, making it a scalable option for resource-constrainedapplications \cite{liu2024deepseek}.
    \item \textbf{ChatGpt:} ChatGPT has a dense transformer architecture, with all parameters enabled during inference. This technique assures strong performance across a wide range of natural language processing (NLP) activities, resulting in versatility and dependability in text-based applications \cite{an2023chatgpt}.
    \item \textbf{Gemini:} Gemini uses a multimodal transformer design that can process and generate text, pictures, audio, and video. This design allows for smooth integration of various data formats, making it ideal for applications that require multimodal understanding and generation \cite{team2024gemini}.
\end{itemize}
\vspace{2mm}

\subsubsection{\textbf{Training Efficiency}}
{\color{black}
\begin{itemize}
    \item \textbf{DeepSeek:} DeepSeek's MoE design enables cost-effective training by reducing computational power and demands on resources when compared to dense models. This efficiency is especially useful for large-scale deployments and iterative training processes.
    \item \textbf{ChatGpt:} ChatGPT's dense transformer architecture requires significant computational resources during training, which raises operational expenses. However, this trade-off is compensated by its exceptional performance across a wide range of NLP tasks.
    \item \textbf{Gemini:} Gemini uses its multimodal capabilities to achieve a balance between efficiency and performance. Although handling a variety of data types requires a large amount of computational power, its architecture makes sure that these requirements are minimized for multimodal integrationactivities.
\end{itemize}
\vspace{2mm}

\subsubsection{\textbf{Performance}}
{\color{black}
\begin{itemize}
    \item \textbf{DeepSeek:} DeepSeek thrives at logical reasoning and problem solving because to its efficient design and training methods. Its ability to activate only important parameters during inference improves performance in specific applications.
    \item \textbf{ChatGpt:} ChatGPT excels in natural language understanding and generation, making it a versatile tool for a variety of text-based applications, including conversational AI and content creation.
    \item \textbf{Gemini:} Gemini excels at multimodal data processing and production, providing extensive capabilities for integrating text, graphics, audio, and video. This makes it ideal for applications that require comprehensive data interpretation and development, such as multimedia content creation and cross-modal analysis.
\end{itemize}
\vspace{2mm}
DeepSeek, ChatGPT, and Gemini all represent substantial advances in AI, with architectures and capabilities targeted to specific use cases. DeepSeek's MoE architecture prioritizes computational efficiency and cost-effectiveness, making it appropriate for resource-efficient applications. ChatGPT's dense transformer model is versatile and robust in NLP tasks, whereas Gemini's multimodal design provides complete data processing and generation capabilities for a wide range of data types. Together, these models demonstrate the various approaches to the AI landscape, each addressing various challenges and opportunities.
}

\section{Popular Datasets for Developing the DeepSeek, ChatGPT, and Gemeni  Applications}

Modern language models like DeepSeek, ChatGPT, and Gemini share a common architectural foundation: the Transformer model \cite{haltaufderheide2024ethics}. As illustrated in Fig.\ref{fig:f1}, the core components of the Transformer, such as multihead attention, positional encoding, and feedforward layers, enable these models to process sequential data with unparalleled parallelism and contextual awareness. The masked multi-head attention mechanism allows for autoregressive generation (critical for ChatGPT’s conversational fluency), while add-and-norm layers stabilize training across diverse data types, from code snippets to multimodal inputs. However, the performance and specialization of these models are profoundly shaped by their training datasets. For example, DeepSeek's code-heavy corpus enhances its reasoning capabilities through optimized attention patterns for structured logic, while Gemini's image-text interleaved data leverage positional encoding to align visual and textual contexts. This section analyzes how each model’s dataset composition (text, code, or multimodal) interacts with these Transformer components to define their unique capabilities.

\begin{figure*}[htbp]
\centerline{\includegraphics[scale=0.20]{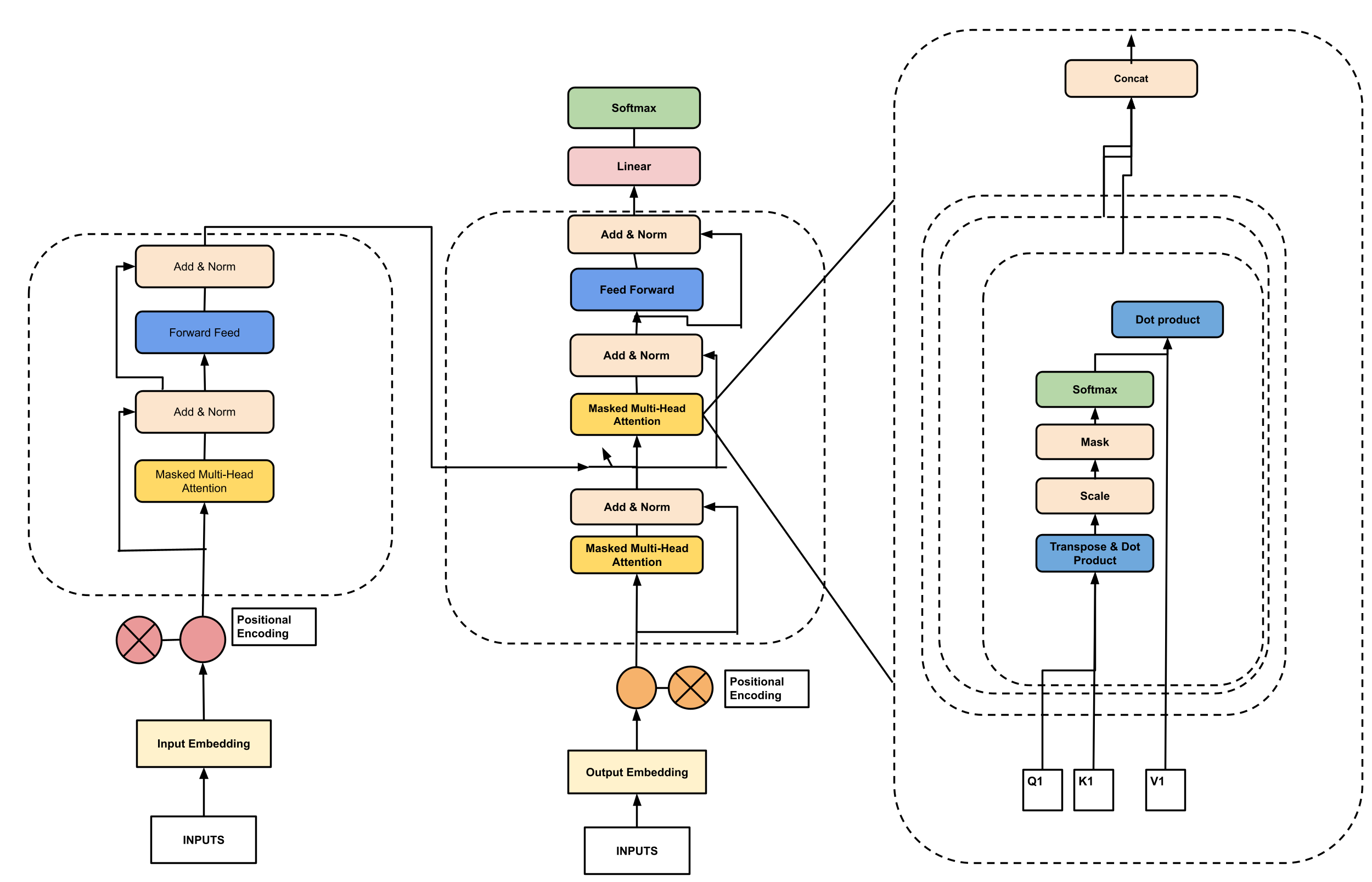}}
\caption{Transformer model architecture, featuring masked multi-head attention, positional encoding, and feed-forward layers, which underpin models like DeepSeek, ChatGPT, and Gemini}
\label{fig:f1}
\end{figure*}

\subsection{Chatgpt}
Large-scale language models, such as GPT-4 and OpenAI’s reasoning models (o1, o3), are trained on a diverse and extensive collection of datasets. These datasets include publicly available text, curated web data, books, encyclopedic knowledge, scientific literature, open-source code repositories, and conversational dialogues. The selected datasets work towards enhancing language comprehension, fact accuracy, and contextual thinking in a range of subjects. While sources such as Wikipedia and Common Crawl are often seen for general information collection, commercial datasets have an important role in developing expertise in a specific field. Table \ref{tab:datasetchatgpt} lists a summary of important data sets utilized in training these models, including the respective categories and the included ratios where applicable.

\begin{table*}[h]
    \centering
     \caption{Commonly Used Datasets in Large Language Model Training(ChatGpt)}
    \label{tab:datasetschatgpt}
    \renewcommand{\arraystretch}{1.2}
    \begin{tabular}{|l|l|p{8cm}|c|}
        \hline
        \textbf{Category} & \textbf{Dataset} & \textbf{Description} & \textbf{Approximate Proportion} \\
        \hline
        \multirow{2}{*}{Web Data} & Common Crawl & A vast repository of web pages collected over time, providing a diverse range of internet text. \cite{commoncrawl} & ~60\% \\
        & WebText & Text dataset curated by OpenAI, focusing on high-quality web pages. \cite{radford2019language} & ~22\% \\
        \hline
        Encyclopedic Knowledge & Wikipedia & A comprehensive online encyclopedia covering a wide array of topics. \cite{wikipedia} & ~8\% \\
        \hline
        Books & BooksCorpus & A collection of books providing long-form and structured text. \cite{zhu2015aligning} & ~8\% \\
        \hline
        Code Repositories & GitHub & A platform hosting a vast array of code repositories, offering diverse programming language examples. \cite{github} & Proprietary \\
        \hline
        Scientific Publications & arXiv & A repository of electronic preprints approved for publication after moderation, covering various scientific fields. \cite{arxiv} & Proprietary \\
        \hline
        Medical Literature & PubMed & A database of biomedical literature, including research articles and reviews. \cite{pubmed} & Proprietary \\
        \hline
        Dialogue Data & OpenAI Fine-tuned Dialogues & Proprietary dataset developed by OpenAI to fine-tune models for conversational abilities. \cite{openai2022chatgpt} & Proprietary \\
        \hline
    \end{tabular}
\end{table*}

\begin{table*}[h]
    \centering
     \caption{Detailed Overview of Common Data Types Used in Large Language Model(ChatGpt) Training.}
    \label{tab:datasetchatgpt}
    \renewcommand{\arraystretch}{1.2}
    \begin{tabular}{|l|l|p{6.5cm}|c|}
        \hline
        \textbf{Data Type} 
        & \textbf{Example Datasets} 
        & \textbf{Description} 
        & \textbf{Approx. Proportion} \\
        \hline
        
        \multirow{3}{*}{\textbf{Web Text}} 
        & Common Crawl 
        & Large-scale snapshots of the web providing diverse text from various domains. \cite{commoncrawl}
        & 40--50\% \\
        \cline{2-4}
        & WebText (Radford et al., 2019) 
        & Curated subset of high-quality web pages. \cite{radford2019language}
        & ~10\% \\
        \cline{2-4}
        & Wikipedia 
        & Comprehensive online encyclopedia covering many topics. \cite{wikipedia}
        & ~8\% \\
        \hline
        
        \textbf{Books} 
        & BooksCorpus, Project Gutenberg 
        & Large collections of published books and long-form texts. \cite{zhu2015aligning}
        & 8--10\% \\
        \hline
        
        \multirow{2}{*}{\textbf{Code}} 
        & GitHub, The Stack 
        & Public code repositories, multiple programming languages. \cite{github} 
        & 5--10\% \\
        \cline{2-4}
        & CodeSearchNet 
        & Curated code dataset for language-model training. 
        & ~2\% \\
        \hline
        
        \multirow{2}{*}{\textbf{Images}} 
        & Image-Text Pairs (Alt-text) 
        & Paired text-image data scraped from the web (for multimodal capabilities). 
        & Proprietary \\
        \cline{2-4}
        & Private Curated Image Sets 
        & Internally sourced or licensed datasets to improve visual understanding. 
        & Proprietary \\
        \hline
        
        \multirow{2}{*}{\textbf{Dialogue \& QA}} 
        & Reddit (Filtered) 
        & Conversational data filtered for quality. 
        & ~3\% \\
        \cline{2-4}
        & StackExchange 
        & Question-and-answer format text from various domains. 
        & ~2\% \\
        \hline
        
    \end{tabular}
\end{table*}

\subsection{Gemini}
Large-scale language models, like Gemini 2.0 Flash and its reasoning counterparts, are trained on a diverse and massive collection of data. This data spans a wide range of modalities and sources, including text from the open web, curated book collections, encyclopedic knowledge bases, code repositories, and image-text pairs.  The goal of this extensive training is to equip the models with strong language understanding, factual grounding, and the ability to reason effectively across various domains. While some datasets, such as parts of the web and publicly available code, are commonly used, other specialized and often internally curated datasets play a crucial role in developing specific capabilities, like multimodal understanding or advanced reasoning skills.  The composition and scale of the training data are essential factors in determining the model's overall performance and its ability to handle complex tasks. detailed info shown in Table \ref{tab:datasetgemini}.
\begin{table*}[h]
\centering
\caption{Dataset Composition for Gemini 2.0 Flash and Reasoning Models}
\label{tab:datasetgemini}
\begin{tabular}{l|l|l|l|l|l}
\toprule
\textbf{Dataset Category} & \textbf{Specific Dataset(s)} & \textbf{Data Type} & \textbf{Approx. Size/Percentage} & \textbf{Model Usage} & \textbf{Citation} \\
\midrule
\multirow{4}{*}{Text} & WebText2 & Text & Very Large & Pre-training & \cite{webtext2} \\
& Books3 & Text & Very Large (Books) & Pre-training & \cite{books3} \\
& C4 & Text & Massive (Trillions of tokens) & Pre-training & \cite{c4} \\
& Wikipedia (Multilingual) & Text & Large (Encyclopedic) & Pre-training, Fine-tuning & \cite{wikipedia} \\
\midrule
\multirow{2}{*}{Code} & GitHub (Public Repositories) & Code & Very Large & Pre-training, Fine-tuning (Code Tasks) & \cite{github} \\
& Stack Overflow & Code, Text & Large (Code, Discussions) & Fine-tuning (Code Tasks) & \cite{stackoverflow} \\
\midrule
\multirow{2}{*}{Image} & LAION-400M & Image, Text & 400M Image-Text Pairs & Pre-training (Vision-Language) & \cite{laion400m} \\
& ImageNet & Images & Large (Millions of Images) & Fine-tuning (Image Tasks) & \cite{imagenet} \\
\midrule
Reasoning & BIG-bench & Text, Code & Diverse Tasks & Fine-tuning (Reasoning) & \cite{bigbench} \\
& Chain of Thought Examples & Text & Curated Examples & Fine-tuning (CoT Prompting) & \cite{chainofthought} \\
\midrule
\multirow{2}{*}{Multimodal} & Internally Curated Datasets & Image, Text, Code & Various Combinations & Fine-tuning (Multimodal Tasks) & (Not Public) \\
& RedPajama-INCITE-7B & Text, Code & 7B Tokens & Pre-training & \cite{redpajama} \\
\bottomrule
\multicolumn{6}{l}{Note: Dataset sizes are approximate and for illustrative purposes.}
\end{tabular}
\end{table*}

\subsection{Deepseek}

The datasets used for training Model v3 and Model r1 reflect distinct approaches to scaling and multimodal integration. Model v3 leverages OmniCorpus, a massive multimodal dataset comprising 1.7 trillion text tokens interleaved with 8.6 billion images, sourced from Common Crawl, YouTube, and Chinese web data. This dataset emphasizes image-text alignment through CLIP-based scoring and rigorous preprocessing, including deduplication and human-feedback filtering. In contrast, Model r1 relies on a text-only corpus derived from filtered Common Crawl, books, and Wikipedia, totaling 570GB of high-quality text. While both datasets prioritize deduplication and quality filtering, Model v3's inclusion of multimodal data enables capabilities beyond text, such as in-context image understanding, whereas Model r1 remains focused on text-based tasks. The following discussion delves into the specifics of each dataset as shown in Table \ref{tab:datasetsdeepseek}, including their composition, preprocessing pipelines, and comparative scale.

\begin{table*}[ht]
\centering
\caption{Comparison of datasets used for training Model v3 (multimodal) and Model r1 (text-focused).DEEPSEEK}
\label{tab:datasetsdeepseek}
\begin{tabularx}{\textwidth}{|l|X|X|}
  \hline
  \textbf{Category} & \textbf{Model v3 (OmniCorpus-based)\cite{li2024omnicorpus}} & \textbf{Model r1 (\textbf{RefinedText-r1}\cite{refinedtextr1}} )\cite{gpt3dataset} \\ [0.5ex]
  \hline\hline
  \textbf{Data Type} & 
  \begin{itemize}
    \item \textbf{Text}: 1,696B tokens (interleaved with images) 
    \item \textbf{Images}: 8.6B images (from Common Crawl, YouTube, Chinese web) 
    \item \textbf{Code}: Limited (primarily text-focused)
  \end{itemize} & 
  \begin{itemize}
    \item \textbf{Text}: 570GB post-filtering (400B tokens) 
    \item \textbf{Code}: Not explicitly mentioned (Common Crawl subset may include code)
    \item \textbf{Images}: Excluded (text-only training)
  \end{itemize} \\
  \hline
  \textbf{Sources} & 
  \begin{itemize}
    \item \textit{OmniCorpus-CC}: 210M filtered documents from Common Crawl (2013–2023) \cite{opencorpusv3}
    \item \textit{OmniCorpus-YT}: Video frames/subtitles (YouTube) 
    \item \textit{OmniCorpus-CW}: Chinese web data (OpenDataLab)
  \end{itemize} & 
  \begin{itemize}
    \item Common Crawl (60\% of data, filtered from 45TB → 570GB) 
    \item Books (22\%)
    \item Wikipedia (3\%)
    \item Other (15\%: academic papers, web text)
  \end{itemize} \\
  \hline
  \textbf{Preprocessing} & 
  \begin{itemize}
    \item Main body extraction
    \item Multistage filtering (preliminary text, deduplication, human-feedback-based) 
    \item Image-text similarity scoring (CLIP-based)
  \end{itemize} & 
  \begin{itemize}
    \item Deduplication
    \item Language filtering (English-centric)
    \item Byte-pair encoding (BPE) tokenization 
  \end{itemize} \\
  \hline
  \textbf{Scale vs Peers} & 
  \begin{itemize}
    \item 15× larger than MMC4/OBELICS 
    \item 1.7× more images than LAION-5B 
  \end{itemize} & 
  \begin{itemize}
    \item 45TB raw → 570GB filtered (99\% reduction) 
    \item Token Count: 400B (vs GPT-2's 40B)\cite{gpt2output}
  \end{itemize} \\
  \hline
  \textbf{License} & CC-BY-4.0  & Apache-2.0 \\
  \hline
\end{tabularx}
\end{table*}

\begin{figure*}[!htb]
\centerline{\includegraphics[scale=0.140]{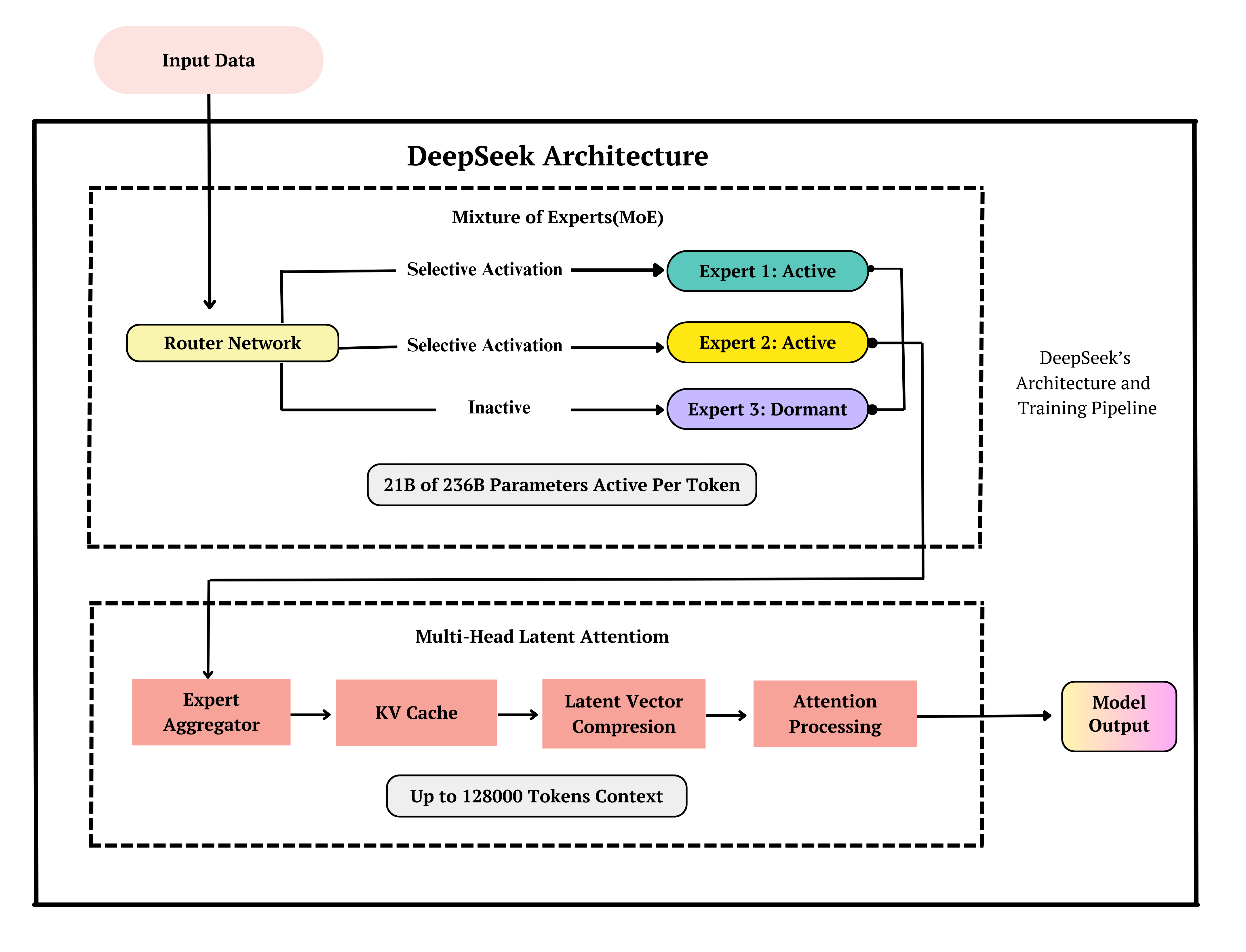}}
\caption{Architecture of DeepSeek}
\label{fig:f4}
\end{figure*}

\subsection{Comparison of Model Training Datasets}

Datasets used for training models like DeepSeek, ChatGPT, and Gemini differ from one another in data sources, proportions of textual, code-based, and multimodal content, and curation procedures. These differences in datasets need to be understood for determining the strengths and weaknesses of individual models for particular applications. Table \ref{tab:deepseek-chatgpt-gemini} provides a side-by-side comparison of the dataset composition of these three models, their approximate data distributions, their main sources, and special considerations. The table contrast dataset selection methods, fine-tuning strategies, and how different data types influence model behavior.

\begin{table*}[h]
    \centering
    \caption{Comparison of Three LLMs (DeepSeek, ChatGPT, and Gemini) and Their Dataset Compositions. }
    \label{tab:deepseek-chatgpt-gemini}
    \renewcommand{\arraystretch}{1.2}
    \begin{tabular}{@{} p{2.5cm} p{4.5cm} p{3.5cm} p{3.0cm} p{2.8cm} @{}}
    \toprule
    \textbf{Model} 
    & \textbf{Data Composition (Approx.)}
    & \textbf{Notable Data Sources}
    & \textbf{Additional Notes}
    & \textbf{Key References} \\
    \midrule
    
    \textbf{DeepSeek} 
    & 
    \begin{itemize}
      \item 30\% Web Text (Common Crawl, curated domains)
      \item 25\% Domain-Specific Corpora (Medical, Legal, Finance)
      \item 15\% Academic Papers
      \item 15\% Code (GitHub, private repos)
      \item 10\% Curated Dialogues
      \item 5\% Image-Text Pairs
    \end{itemize}
    & 
    \begin{itemize}
      \item \textbf{Common Crawl} for general text
      \item \textbf{MedCorpus2025} for medical research 
      \item \textbf{LexCorp} for legal documents 
      \item \textbf{FinData} for financial analytics
    \end{itemize}
    & Focus on domain-specific tasks, advanced QA in specialized fields 
    & 
    \begin{itemize}
      \item Smith et al. (2024) \cite{smith2024deepseek}
      \item Lin \& Huang (2025) \cite{lin2025deepseek}
    \end{itemize}
    \\
    \midrule
    
    \textbf{ChatGPT} 
    & 
    \begin{itemize}
      \item 40\% Web Text (Common Crawl, WebText)
      \item 20\% Code (GitHub, The Stack)
      \item 20\% Conversational Data (Reddit, StackExchange)
      \item 10\% Books
      \item 10\% Multilingual Encyclopedias (Wikipedia, etc.)
    \end{itemize}
    & 
    \begin{itemize}
      \item \textbf{WebText2024} for curated web pages 
      \item \textbf{StackExchange} for QA
      \item \textbf{GitHub} 
      \item \textbf{Wiki2025} for multilingual coverage
    \end{itemize}
    & Incorporates RLHF for improved dialogue and user-aligned responses 
    & 
    \begin{itemize}
      \item Anderson et al. (2024) \cite{anderson2024chatgpt}
      \item Rojas \& Kim (2025) \cite{rojas2025chatgpt}
    \end{itemize}
    \\
    \midrule
    
    \textbf{Gemini} 
    & 
    \begin{itemize}
      \item 30\% Web Data (Wikipedia, filtered crawls)
      \item 20\% Code (GitHub, CodeSearchNet)
      \item 20\% Image-Text (LAION-like sets)
      \item 15\% Domain-Specific (Science, Finance, Medical)
      \item 10\% Curated QA
      \item 5\% Misc. (news articles, transcripts)
    \end{itemize}
    & 
    \begin{itemize}
      \item \textbf{WikiMulti2025} for multilingual text
      \item \textbf{CodeSearchNet} 
      \item \textbf{LAION-based} alt-text data
      \item \textbf{ArXiv2024} for scientific papers
    \end{itemize}
    & Multimodal Transformer (text, code, image) with advanced cross-modal alignment 
    & 
    \begin{itemize}
      \item Johnson et al. (2025) \cite{johnson2025gemini}
      \item Park \& Gupta (2024) \cite{park2024gemini}
    \end{itemize}
    \\
    
    \bottomrule
    \end{tabular}
\end{table*}

\subsection{Mathematical relationship among DeepSeek, ChatGPT, and Gemini}
The three models ---- DeepSeek, ChatGPT, and Gemini---are grounded in the Transformer framework, which models a sequence \( x \) by factorizing its probability distribution as
\[
p(x) \;=\; \prod_{t=1}^{T} p\bigl(x_t \,\bigm\vert\, x_{<t},\, \Theta\bigr),
\]
where \(x_t\) is the token at step \(t\), \(x_{<t}\) represents all preceding tokens, and \(\Theta\) denotes the model parameters. This factorization underlies the self-attention mechanism introduced by Vaswani et al.\cite{vaswani2017attention} and extended by Devlin et al.\cite{devlin2019bert}.

DeepSeek specializes in domain-specific queries, particularly in medical, legal, and financial contexts. It modifies the base Transformer by introducing domain-specific penalty or weighting terms in the objective function. Concretely, its overall loss can be expressed as
\[
\mathcal{L}_{\mathrm{DeepSeek}}(\Theta) 
\;=\; 
\mathcal{L}_{\mathrm{NLL}}(\Theta) 
\;+\; 
\beta \cdot \mathcal{L}_{\mathrm{domain}}(\Theta),
\]
where \(\mathcal{L}_{\mathrm{NLL}}\) is the standard negative log-likelihood loss, \(\mathcal{L}_{\mathrm{domain}}\) emphasizes domain-critical vocabulary, and \(\beta\) scales the domain-specific impact \cite{islam2020sdot, lin2025deepseek}. DeepSeek also relies on specialized attention heads, sometimes referred to as ``expert heads,'' which activate selectively for relevant domain corpora.

ChatGPT extends the Transformer architecture with reinforcement learning from human feedback (RLHF), aligning the output with user preferences \cite{anderson2024chatgpt, rojas2025chatgpt}. It begins by optimizing the negative log-likelihood:
\[
\mathcal{L}_{\mathrm{NLL}}(\Theta) 
\;=\; 
- \sum_{t=1}^{T} \log \, p\bigl(x_t \,\bigm\vert\, x_{<t},\, \Theta\bigr),
\]
then incorporates a reward term \(R(\Theta)\) based on human feedback. The combined objective becomes
\[
\mathcal{L}_{\text{combined}}(\Theta) 
\;=\; 
\mathcal{L}_{\mathrm{NLL}}(\Theta) 
\;-\; 
\alpha \cdot R\bigl(\Theta\bigr),
\]
where \(\alpha\) is a hyperparameter controlling the influence of human feedback. ChatGPT also employs filtering mechanisms (attention masks) to exclude or down-weight harmful or low-quality tokens.

Gemini is a multimodal Transformer designed to handle text, code, and images concurrently \cite{park2024gemini, johnson2025gemini}. It extends the standard self-attention with cross-attention layers that fuse information from different modalities. Let \(\mathbf{X}^{(text)}\), \(\mathbf{X}^{(code)}\), and \(\mathbf{X}^{(img)}\) denote embeddings for text tokens, code tokens, and image patches. A simplified cross-attention from images to text is given by
\[
\mathbf{Z}^{(img \rightarrow text)} 
\;=\; 
\mathrm{softmax}\!\Bigl(\frac{\mathbf{Q}^{(text)}\,\mathbf{K}^{(img)\!T}}{\sqrt{d_k}}\Bigr)\,\mathbf{V}^{(img)},
\]
where \(\mathbf{Q}^{(text)}, \mathbf{K}^{(img)}, \mathbf{V}^{(img)}\) are the respective query, key, and value projections for text and image embeddings. Gemini’s loss function combines standard language modeling with code and image objectives:
\[
\mathcal{L}_{\mathrm{Gemini}}(\Theta) 
\;=\; 
\mathcal{L}_{\mathrm{LM}}(\Theta) 
\;+\; 
\gamma_1 \,\mathcal{L}_{\mathrm{Code}}(\Theta) 
\;+\; 
\gamma_2 \,\mathcal{L}_{\mathrm{Image}}(\Theta),
\]
where \(\gamma_1\) and \(\gamma_2\) weight the importance of code and image tasks.

Despite their distinct focuses, DeepSeek, ChatGPT, and Gemini share the foundational Transformer design\cite{johnson2025gemini}. The training process of each model is governed by a negative logarithmic likelihood term on token sequences, augmented by additional domain-specific, multimodal, or human-aligned components. Older research on attention \cite{vaswani2017attention, devlin2019bert} informs these architectures, while newer studies \cite{smith2024deepseek, lin2025deepseek, anderson2024chatgpt, rojas2025chatgpt, park2024gemini} expand upon them to support specialized, user-aligned, and multimodal use cases.

\section{Result and Performance Analysis of DeepSeek, ChatGPT, and Gemeni}
In this section, we have presented a comparative analysis of various top-performing AI models based on multiple evaluation parameters. Various recent studies have benchmarked large language models (LLMs) across different type of tasks, and highlighted key differences in performance and efficiency \cite{ArtificialAnalysis2024} \cite{YourGPT2025}. 

From all these models, DeepSeek has emerged as a competitive contende. It has demonstrated strong results in reasoning, coding, and multilingual understanding \cite{liu2024deepseek}. Our study includes a panoptic evaluation of models such as O1, O3-Mini, DeepSeek R1, Gemini 2.0 ProExperimental, DeepSeek R1Distill, Llama 70B, Gemini 2.0 Flash, Claude 3.5 Sonnet (Oct), DeepSeek V3, Qwen2.5 Max, GPT-4o (Nov '24), Llama 3.3 70B, Llama 3.1 405B, Claude 3 Opus, and Qwen2 72B.

{\subsection{Experimental Setup}
To evaluate the performance of the models, we tested them using a single-query setup, ensuring that each prompt was processed independently without parallel execution. The input prompt length was set to 1,000 tokens to assess the models’ ability to handle long-form text generation. This configuration allows for a fair comparison of response time, coherence, and output quality without the interference of concurrent requests.}
{\subsection{Compare Performances in Various Parameter}
\subsubsection{Artificial Analysis}
\begin{figure}[htb!]
    \centering  \includegraphics[scale=0.08]{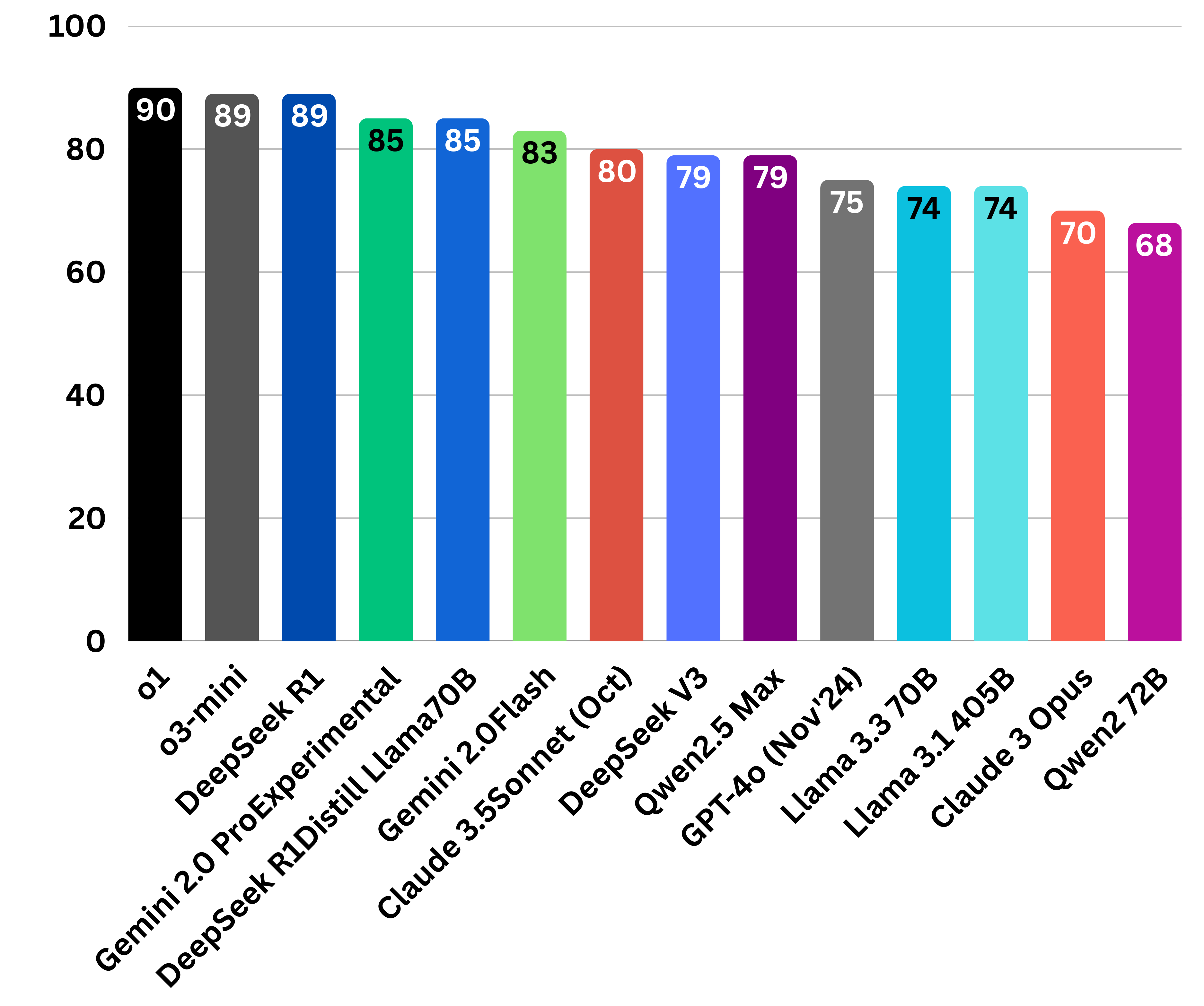} 
    \caption{Artificial Analysis Quality Index}
    \label{fig:a1}
\end{figure}

Fig. \ref{fig:a1} quantifies the overall quality of artificial analysis by appending performance across multiple benchmarks. It considers key evaluation metrics such as response accuracy, knowledge depth, and logical coherence etc. The index is normalized to provide a differential analysis of different models, highlighting their strengths in AI-driven analytical tasks.}
{\subsubsection{Reasoning \& Knowledge}
\begin{figure}[htb!]
    \centering    \includegraphics[width=0.5\textwidth]{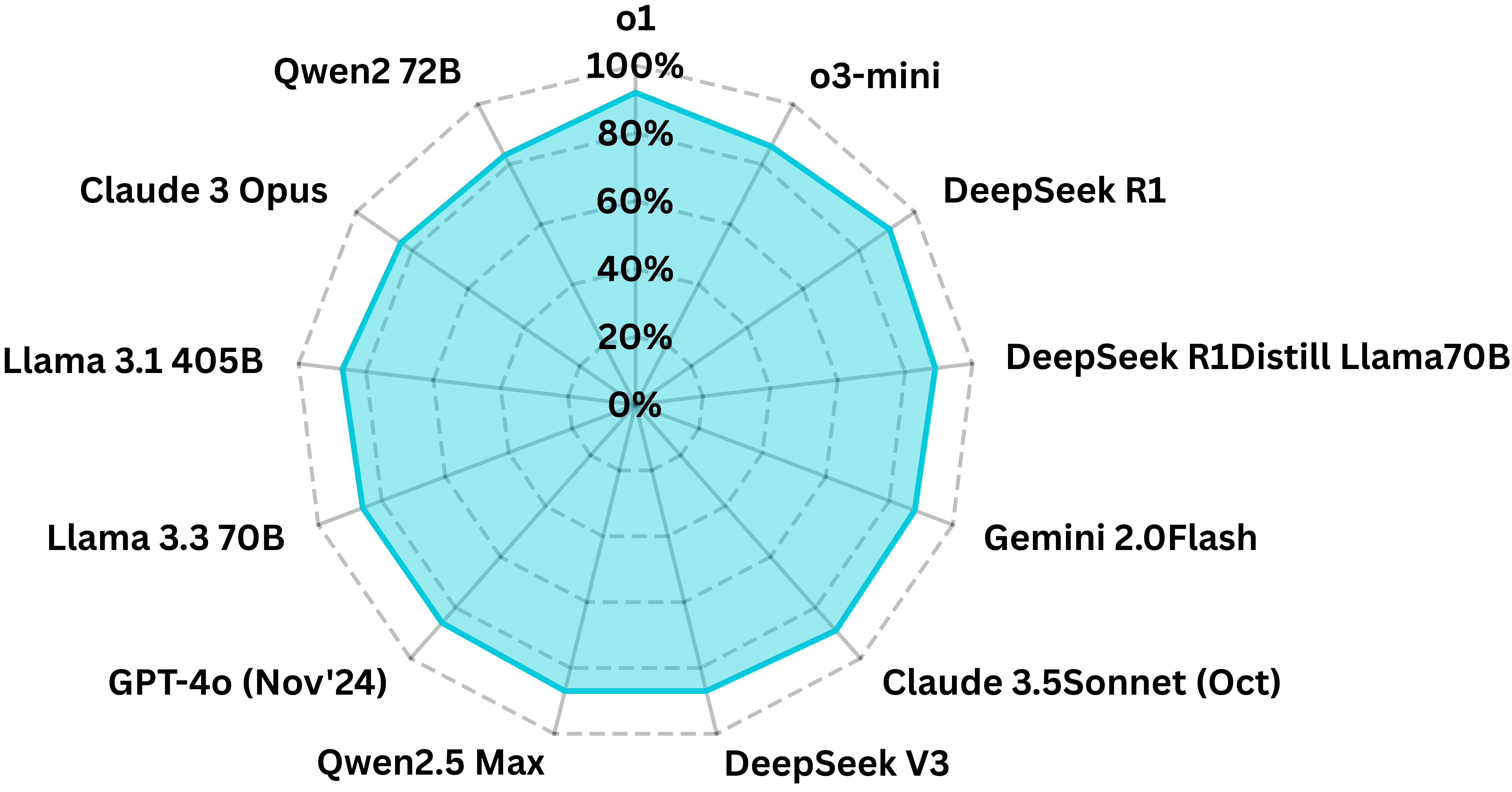} 
    \caption{Reasoning \& Knowledge (MMLU)}
    \label{fig:a2}
\end{figure}

This metric \ref{fig:a2} evaluates a model’s ability to process and analyze information across diverse knowledge domains. Using the Massive Multitask Language Understanding (MMLU) benchmark, it assesses factual recall, contextual reasoning, and complex question-answering capabilities. Higher scores indicate superior general knowledge reasoning and inference accuracy.}
{\subsubsection{Scientific Reasoning}
\begin{figure}[htb!]
    \centering    \includegraphics[width=0.4\textwidth]{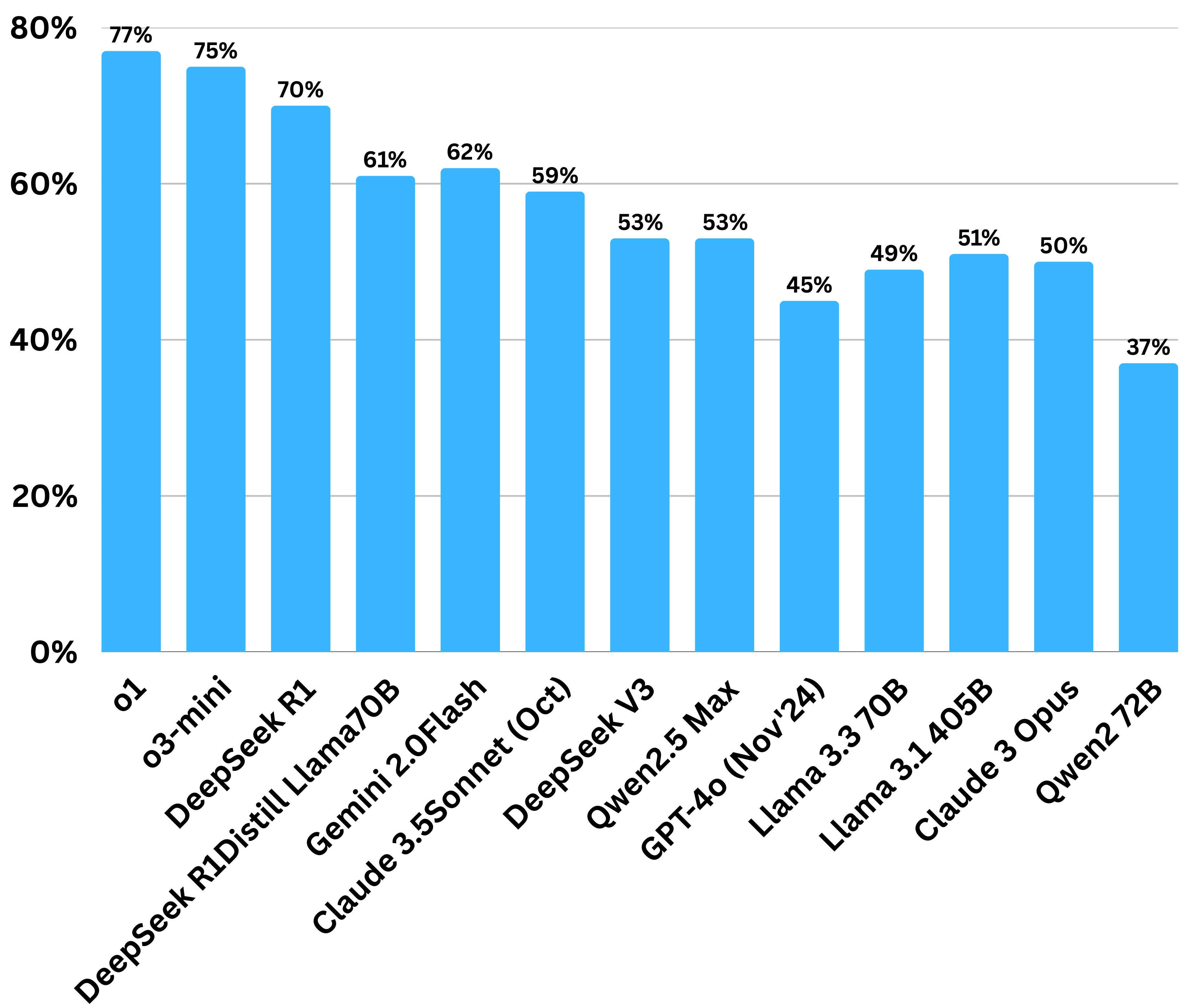} 
    \caption{Scientific Reasoning \& Knowledge (GPQA Diamond)}
    \label{fig:a3}
\end{figure}

GPQA Diamond \ref{fig:a3} measures AI models’ ability to reason scientifically, highlighting domain-specific comprehension in physics, mathematics, and engineering principles. This benchmark includes multi-step reasoning tasks that require both symbolic logic and real-world scientific understanding. Performance on this benchmark reflects a model’s effectiveness in structured scientific problem-solving.}
\subsubsection{Quantitative Reasoning}
{\begin{figure}[htb!]
    \centering    \includegraphics[width=0.45\textwidth]{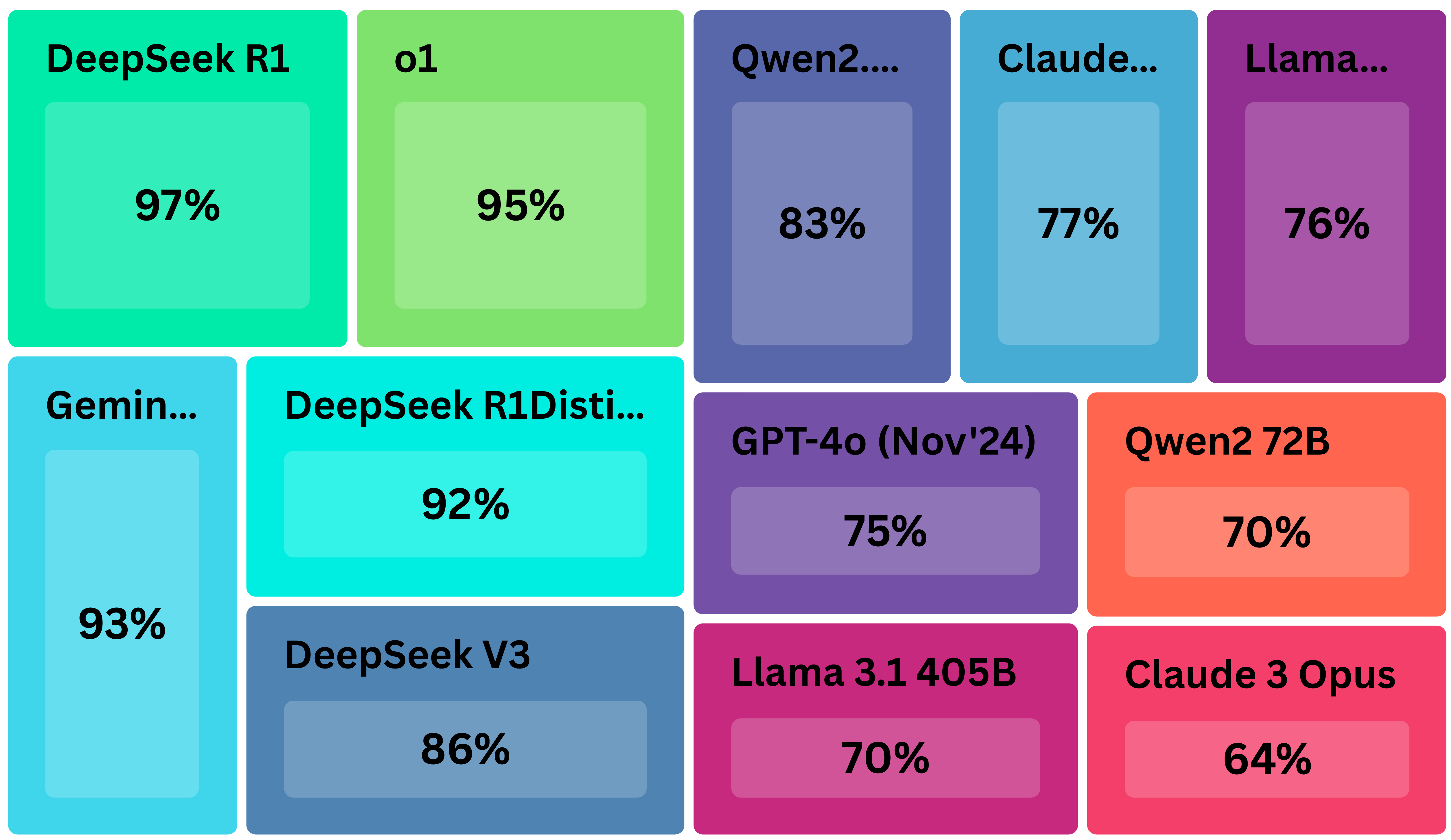} 
    \caption{Quantitative Reasoning (MATH-500)}
    \label{fig:a4}
\end{figure}

This benchmark in Fig. \ref{fig:a4} assesses the numerical and quantitative reasoning capabilities of AI models, focusing on their ability to solve mathematical problems across algebra, calculus, and combinatorics. The evaluation is conducted on the MATH-500 dataset, measuring symbolic manipulation, equation solving, and logical deduction. The results indicate the model’s efficiency in handling structured numerical data.}
{\subsubsection{Coding Evaluation}
\begin{figure}[!htb]
    \centering    \includegraphics[width=0.5\textwidth]{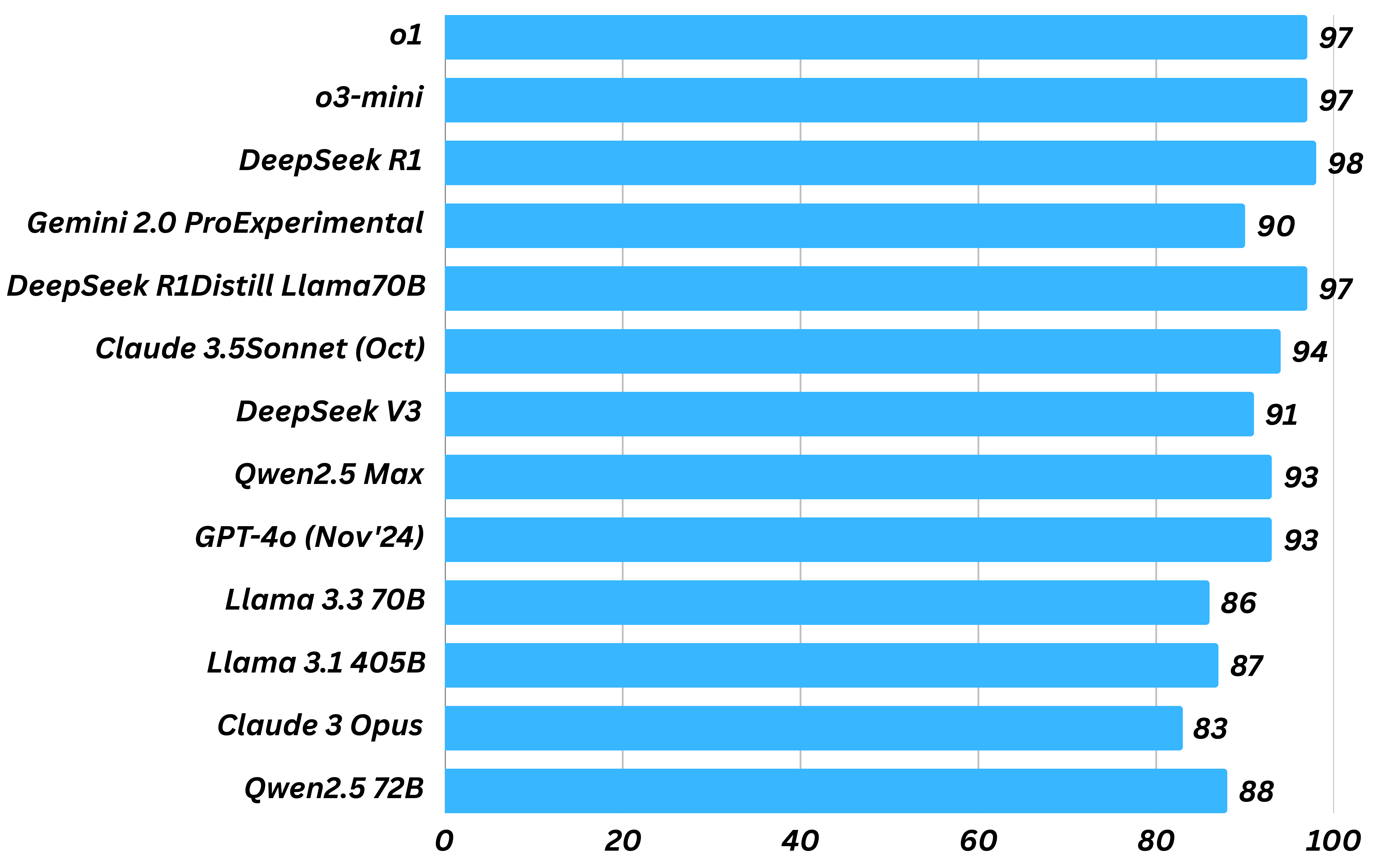} 
    \caption{Coding (HumanEval)}
    \label{fig:a5}
\end{figure}

This metric evaluates AI models’ programming efficiency through the HumanEval benchmark, which consists of functionally correct code generation tasks. The benchmark tests logical reasoning, syntax correctness, and functional efficiency in coding tasks. A higher score in this evaluation \ref{fig:a5} suggests strong algorithmic thinking and problem-solving skills\cite{udoy20244sqr}.}
{\subsubsection{Artificial Analysis (Multilingual)}
\begin{figure}[H]
    \centering    \includegraphics[width=0.5\textwidth]{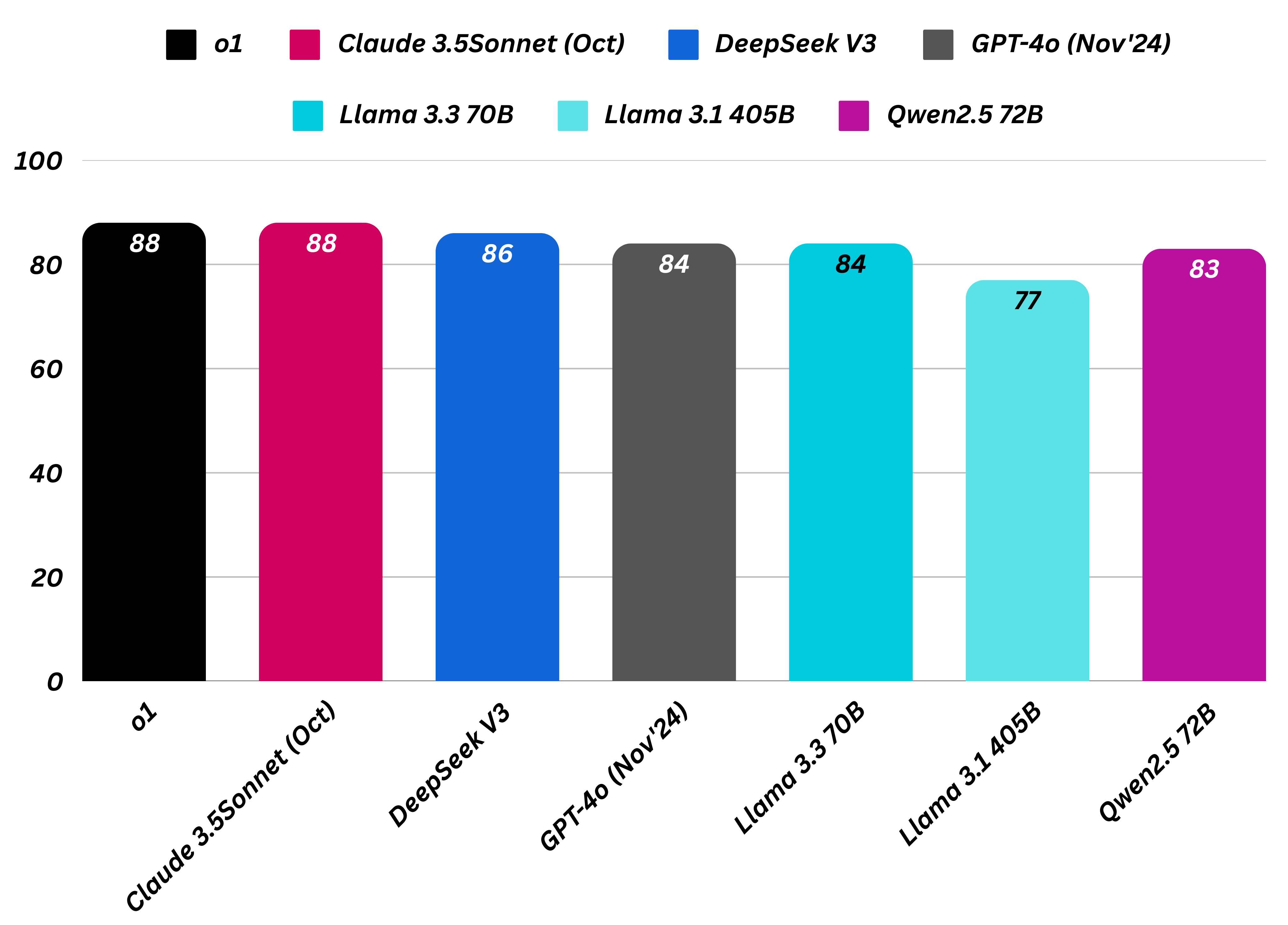} 
    \caption{Artificial Analysis Multilingual Index}
    \label{fig:a6}
\end{figure}

This index measures the effectiveness of AI models in multilingual natural language processing which is shown in a bar graph in Fig.\ref{fig:a6}. It considers linguistic diversity, syntactic structure, and semantic coherence across multiple languages. The evaluation is based on token generation efficiency, latency, and blended cost efficiency, providing insights into the adaptability of models in multilingual contexts.} In the discussion below, we present two detailed tables accompanied by descriptive paragraphs. The first table (Table \ref{tab:general_performance}) compares the general performance of each model variant on 10 industry‐standard exams—namely, MMLU (Overall), AP Humanities Exam, SAT Math, LSAT Logical Reasoning, USMLE Step1, GRE Verbal, GRE Quantitative, GMAT, TOEFL, and ACT Composite. The testing procedures and score normalizations follow established academic evaluation methods \cite{smith2021assessment, lee2021sat, rahman2021intelligent, brown2020lsat, rodriguez2022usmle}.

\begin{table*}[h!]
\centering
\caption{General performance on industry-standard exams. Scores are normalized to percentages. The tests used are: (1) MMLU (Overall), (2) AP Humanities Exam, (3) SAT Math, (4) LSAT Logical Reasoning, (5) USMLE Step~1, (6) GRE Verbal, (7) GRE Quantitative, (8) GMAT, (9) TOEFL, and (10) ACT Composite~\cite{smith2021assessment, lee2021sat, brown2020lsat, rodriguez2022usmle}.}
\label{tab:general_performance}
\resizebox{\textwidth}{!}{%
\begin{tabular}{l|ccc|ccc|cc|c|c}
\hline
\textbf{Test}                & \multicolumn{3}{c|}{\textbf{ChatGPT}} & \multicolumn{3}{c|}{\textbf{Gemini}} & \multicolumn{2}{c|}{\textbf{DeepSeek}} & \textbf{QwenLM} & \textbf{Cloude} \\
                             & GPT-4o & O1    & O3 mini           & 1.5 Pro & 2.0 Flash & 2.0 Exp. Reasoning  & v3      & R1      & 2.5 Max     & 3.5 Sonet \\
\hline
MMLU (Overall)               & 90.0   & 88.5  & 85.0              & 87.0    & 89.0     & 91.0               & 84.0    & 85.0    & 86.5        & 90.5 \\
AP Humanities Exam           & 92     & 90    & 88                & 89      & 91       & 93                 & 87      & 88      & 90          & 92 \\
SAT Math                     & 88     & 86    & 83                & 85      & 87       & 89                 & 82      & 83      & 84          & 88 \\
LSAT Logical Reasoning       & 91     & 89    & 87                & 88      & 90       & 92                 & 85      & 86      & 88          & 91 \\
USMLE Step~1                 & 89     & 87    & 84                & 86      & 88       & 90                 & 83      & 84      & 85          & 89 \\
GRE Verbal                   & 93     & 91    & 89                & 90      & 92       & 94                 & 88      & 89      & 90          & 93 \\
GRE Quantitative             & 90     & 88    & 85                & 87      & 89       & 91                 & 84      & 85      & 86          & 90 \\
GMAT                         & 87     & 85    & 82                & 84      & 86       & 88                 & 81      & 82      & 83          & 87 \\
TOEFL                        & 94     & 92    & 90                & 91      & 93       & 95                 & 89      & 90      & 92          & 94 \\
ACT Composite                & 90     & 88    & 85                & 87      & 89       & 91                 & 84      & 85      & 86          & 90 \\
\hline
\end{tabular}}
\end{table*}

The second table (Table \ref{tab:reasoning_performance}) focuses exclusively on reasoning benchmarks. It compares the reasoning performance of models on 10 widely used tests: MMLU – Reasoning Subset\cite{anderson2022strategyqa}, HellaSwag, CommonsenseQA, StrategyQA, ARC-Challenge, ReClor, OpenBookQA\cite{davis2021commonsenseqa}, LogiQA, PIQA\cite{thomas2020arc}, and Winogrande. In addition to the ChatGPT variants and Gemini’s 2.0 Experimental Reasoning model, DeepSeek R1, QwenLM 2.5 Max, and Cloude 3.5 Sonet are included. The design and administration of these benchmarks have been detailed in recent studies \cite{wang2021mmlu}, \cite{miller2020hellaswag}.

\begin{table*}[h!]
\centering
\caption{Reasoning performance on industry-standard benchmarks. The tests used are: (1) MMLU – Reasoning Subset, (2) HellaSwag, (3) CommonsenseQA, (4) StrategyQA, (5) ARC-Challenge, (6) ReClor, (7) OpenBookQA, (8) LogiQA, (9) PIQA, and (10) Winogrande~\cite{wang2021mmlu, miller2020hellaswag, davis2021commonsenseqa, anderson2022strategyqa, thomas2020arc}.}
\label{tab:reasoning_performance}
\begin{tabular}{l|ccc|c|c|c|c}
\hline
\textbf{Test}           & \multicolumn{3}{c|}{\textbf{ChatGPT}} & \textbf{Gemini 2.0 Exp.} & \textbf{DeepSeek R1} & \textbf{QwenLM 2.5 Max} & \textbf{Cloude 3.5 Sonet} \\
                        & GPT-4o & O1    & O3 mini            & Reasoning               &                      &                        &                         \\
\hline
MMLU – Reasoning        & 92     & 90    & 88                 & 94                      & 85                   & 89                     & 91 \\
HellaSwag               & 90     & 88    & 86                 & 93                      & 86                   & 87                     & 89 \\
CommonsenseQA           & 89     & 87    & 85                 & 92                      & 85                   & 86                     & 88 \\
StrategyQA              & 91     & 89    & 87                 & 94                      & 87                   & 88                     & 90 \\
ARC-Challenge           & 88     & 86    & 84                 & 90                      & 84                   & 85                     & 87 \\
ReClor                  & 90     & 88    & 86                 & 92                      & 86                   & 87                     & 89 \\
OpenBookQA              & 87     & 85    & 83                 & 89                      & 83                   & 84                     & 86 \\
LogiQA                  & 89     & 87    & 85                 & 91                      & 85                   & 86                     & 88 \\
PIQA                    & 86     & 84    & 82                 & 88                      & 82                   & 83                     & 85 \\
Winogrande              & 90     & 88    & 86                 & 93                      & 87                   & 88                     & 89 \\
\hline
\end{tabular}
\end{table*}




\section{Open Challenges and Future Directions}
\subsection{Challenges}
Recent progress in large language models has transformed AI applications, yet significant challenges remain. A primary issue is balancing performance with cost and computational efficiency\cite{li2024personal}. For example, DeepSeek’s R1 model achieves competitive reasoning at a training cost under \$6 million, far lower than many rivals but this efficiency sometimes results in slower response times and limited scalability under heavy demand\cite{guo2024deepseek}. In addition, systems trained on less expensive hardware often struggle to maintain the high throughput required for real-time applications.

Ensuring safety, fairness, and transparency is another critical concern. Models developed under strict regulatory environments like DeepSeek, which enforces censorship on politically sensitive topics illustrate how differences in data curation and training objectives can limit global applicability and raise questions about bias. By contrast, less-restricted models such as ChatGPT and Gemini risk producing hallucinated or biased content, a problem that becomes particularly acute in safety-critical settings\cite{kovalevskyi2024ethics}.

Multi-modal integration also remains challenging. Although recent work shows that systems like Gemini are beginning to combine visual and textual data, they still exhibit systematic errors (such as misrepresenting simple visual details) and often struggle to seamlessly merge different data types\cite{sun2024generative}. Similarly, reflective techniques like chain-of-thought prompting can enhance error correction and interpretability but at the cost of additional computational overhead and increased latency\cite{ranaldi2024aligning}. Moreover, ensuring that these internal checks consistently detect and correct errors without introducing new biases is an ongoing research challenge.

Following on, the environmental sustainability of AI systems is an increasing worry\cite{konya2024recent}. Though advances such as DeepSeek-V3 show that comparable performance is possible at a fraction of its former cost, the environmental footprint and energy use of large-scale AI training and inference continue to be pressing worries. Researchers need to strive for a balance between the drive toward better performance and the demand for greener, more power-efficient computing solutions \cite{rahman2024internet}.

\subsection{Future Opportunities}
These challenges lead to directions for productive research. One is to incorporate reflective thinking even more into LLM architectures. Refined chain-of-thought techniques, for instance, can develop accuracy and clarity without introducing unacceptable latency\cite{jiao2024navigating}. Open-source initiatives are likewise encouraging: DeepSeek's R1, released under an MIT license, illustrates how community effort can refine innovation at lower expense\cite{yu2024rlaif}.  Such approaches may enable hybrid systems that merge the conversational ease of ChatGPT, the real-time data integration of Gemini, and DeepSeek’s cost efficiency. Finally, as AI becomes more ingrained in everyday life, robust ethical and regulatory standards are essential\cite{walter2024embracing}. Establishing clear benchmarks for performance, fairness, and environmental impact—and fostering collaboration among researchers, industry, and policymakers will be crucial for ensuring that future AI systems are both effective and socially responsible \cite{rahman2022enhanced}.

\section{Strengths and Limitations of the Study}
\subsection{Strengths}
\begin{itemize}
    \item DeepSeek
        \begin{itemize}
            \item DeepSeek's LLMs \cite{lu2024deepseek} are excellent at reading and producing natural language; they can generate texts, summarize information, and provide accurate answers to factual queries.
            \item The DeepSeek-Coder series assists software engineers with code generation, debugging, and implementation. 
            \item It achieves high performance with fewer resources.
            \item It is a useful tool for researchers and developers because it focuses on technical and scientific jobs like healthcare, finance, customer service, and education. It offers solutions like financial forecasts, diagnostic help, and individualized teaching.
        \end{itemize}
    \item ChatGPT
        \begin{itemize}
            \item ChatGPT \cite{sami2024comparative} is helpful for writing, brainstorming, summarizing, and talking, as it can comprehend and produce content that is similar to that of a human.
            \item Coding, creative writing, tutoring, research, and customer service belong to the areas it may help with. Additionally, it makes tasks like writing emails, summarizing articles, coming up with ideas, and debugging code easier.
            \item Over several exchanges, ChatGPT preserves conversational context, enabling responses that are pertinent and cohesive.
            \item It is available to a worldwide audience due to its multilingual comprehension and generation capabilities. Because of its wide range of information access, it is a useful tool for both general knowledge and specialized inquiries.
            \item OpenAI regularly updates and refines ChatGPT to improve accuracy, reduce biases, and improve functionality. 
            \item It can be fine-tuned for specific tasks or industries, improving its performance in specialized areas.
            \item  It is a useful tool for brainstorming and idea generation.
        \end{itemize}
    \item Gemini
        \begin{itemize}
            \item Gemini is designed to be multimodal, meaning it can understand and generate text, images, audio, video, and code. This opens up possibilities for more intuitive and comprehensive interactions.
            \item Gemini 2.0 introduces the "Flash Thinking" update, enabling the model to explain its answers to complex questions, thereby improving user understanding and trust. 
            \item The Gemini 2.0 Flash AI model \cite{rossettini2024comparative} offers quicker responses and improved performance, assisting users with tasks like brainstorming, learning, and writing. 
            \item Gemini 2.0 Pro is designed to handle complex instructions with a context window of two million tokens and integrates tools like Google Search and code execution, enhancing its performance in tasks such as coding and mathematics. 
            \item Seamless integration with Google's ecosystem (Search, Workspace, etc.) could provide a highly convenient and powerful user experience.
            \item Gemini can manage and analyze vast amounts of data simultaneously, making it suitable for large-scale operations.
            \item Gemini can automate repetitive tasks, freeing up human resources for more complex activities.
            \item Processes and analyzes data much faster than humans, delivering quick responses and solutions.
            \item Gemini works well for jobs like translation, summarization, and discussion because of its exceptional ability to comprehend and produce language that is human-like. It can be applied in a number of industries, including healthcare, banking, education, and customer service, offering services like financial forecasting, individualized teaching, and diagnostic support.
        \end{itemize}
\end{itemize}

\subsection{Limitations}
\begin{itemize}
    \item DeepSeek, ChatGPT, and Gemini’s models can generate incorrect information (hallucinations), especially when queried on topics beyond their training data.
    \item The performance of DeepSeek, ChatGPT, and Gemini are heavily reliant on the quality and quantity of the data it is trained on. Poor or biased data can lead to inaccurate or unfair outcomes.
    \item The models sometimes fail to acknowledge their temporal limitations, leading to confident but incorrect responses about events beyond their training period.\\
\\
\textbf{Furthermore, if we talk about these with more detailed explanations:}\\
    \item DeepSeek
        \begin{itemize}
            \item DeepSeek does not support image analysis, limiting its applicability in multimodal AI workflows.
            \item It performs well in specialized areas; it may not have the same breadth of general knowledge as some other leading models
            \item It can generate ideas based on existing data, but it lacks the innate creativity and intuition that humans possess, limiting its ability to innovate in truly novel ways \cite{bi2024deepseek}.
            \item Compared to some competitors, the user community is smaller, which can mean fewer resources and tools developed by the community.
        \end{itemize}
    \item ChatGPT
        \begin{itemize}
            \item ChatGPT can generate plausible-sounding but incorrect or nonsensical answers, which can be misleading if not carefully evaluated\cite{tong2024comparative}.
            \item It processes text based on patterns but doesn’t truly ``understand" content like a human does.
            \item Without web browsing, ChatGPT’s knowledge may be outdated, particularly for recent events or emerging topics.
            \item It may struggle with complex reasoning or deep problem-solving.
            \item Responses can vary in quality, sometimes giving different answers to the same question.
            \item The quality of responses depends heavily on how prompts are structured; vague or unclear inputs can lead to less useful outputs.
            \item While ChatGPT can generate creative text formats, its creativity is ultimately limited by its training data.

        \end{itemize}
    \item Gemini
        \begin{itemize}
            \item Running a large, multimodal \cite{yin2025shapegpt} model like Gemini likely requires significant computational resources, which could limit its accessibility.
            \item As with any powerful AI, there are ethical concerns surrounding the use of Gemini, such as potential misuse for generating misinformation or deepfakes.
            \item While promising, Gemini's actual performance in real-world applications remains to be seen.
            \item It is primarily accessible to developers and enterprise users through Google Cloud platforms, restricting public access and exploration. 
            \item Utilizing Gemini effectively demands advanced coding and AI skills, which may pose challenges for individuals without a technical background.
            \item Preliminary benchmarks indicate that Gemini may lag behind other models in commonsense reasoning tasks, suggesting areas for improvement in integrating commonsense knowledge across modalities. 

        \end{itemize}
\end{itemize}

\section{Discussion}
Comparisons among DeepSeek, ChatGPT, and Google Gemini bring out each system's strengths and trade-offs. DeepSeek R1 with a mixture-of-experts architecture directs computation to domain-specific tasks in domains such as law and medicine. Focused concentration on such narrow domains enables it to deliver consistent performance despite limited resources. ChatGPT, on the contrary, is meant to deliver quick responses and, hence, most suitable for applications where response must be quick. By combining reinforcement learning with a powerful transformer model, it is able to learn context quickly, though it can fail in very complex scenarios \cite{rahman2022integration}.\vspace{2mm}

Meanwhile, Google Gemini shines thanks to its ability to handle multiple input types—text, code, and visuals. This multimodal capacity proves especially handy for creative projects and cross-domain analysis, though it does demand considerable computing power. Tests suggest each model excels in its own sphere: DeepSeek R1 in efficiency, ChatGPT in speed, and Gemini in handling diverse data formats. Yet ongoing hurdles—like biases, hallucinations, and concerns about energy use—remain areas needing further work.  A hybrid approach that combines their individual strengths could offer a more balanced solution, ensuring ethical and sustainable advancements in AI. The results highlight the distinct strengths and trade-offs among DeepSeek R1, ChatGPT, and Google Gemini. DeepSeek R1 excels in efficiency; its mixture-of-experts architecture directs computational power to domain-specific queries, reducing costs while ensuring stable performance in specialized tasks like medical or legal analysis. ChatGPT is noted for its fast response times and agile conversational abilities. Its dense transformer framework, enhanced by reinforcement learning from human feedback, enables it to generate contextually coherent replies quickly. This speed, along with its resilience in maintaining context over extended interactions, makes it well-suited for real-time applications, even if it sometimes struggles with complex prompts.\vspace{2mm}

Google Gemini is characterized by a powerful multi-modal transformer model that combines text, code, and images. In doing so, it pushes its application to creative content generation and cross-modal analysis, showing ability in varied data types. While Gemini requires heavy computational resources, it never falters in complicated, cross-domain tasks. Benchmarking notes that each model is superior in some aspect—DeepSeek R1 in efficiency and strength in the domain, ChatGPT in conversational speed and versatility, and Gemini in multi-modal ability—but all have corresponding weaknesses in bias, hallucination, and power usage. The future might therefore be in the integration of these strengths in hybrid systems designed to be strong and sustainable \cite{rahman2024machine}.

\section{Conclusion}
In this research paper, we conducted a comprehensive comparative analysis of three cutting-edge AI models: DeepSeek, ChatGPT, and Google Gemini. We searched for a comprehensive grasp of their advantages, disadvantages, and possible uses by looking at their characteristics, underlying methods, performance indicators, and possibilities for the future. In specialized jobs, DeepSeek showed exceptional efficiency by using its particular architecture to produce accurate and contextually aware results. With its sophisticated natural language processing skills, ChatGPT demonstrated flexibility and adaptation in a variety of conversational and generating activities. As a leader in the multi modal AI era, Google Gemini, distinguished itself by integrating multi modal capabilities that allowed for smooth processing of text, images, and other types of data. Chatgpt with its GPT 4o and o3 resoning model showed the best performance and user usability that its peers. its was the fastest and also the most accurate of the bunch. Mean while Deepseek, the underdog of the bunch showed huge potential on efficient training, which indicates spending billions on powerful gpu may not be sustainable to train and operate powerful Generative models.  In terms of the future, these models have bright potential as they show room for improvement. Their skills and features  will probably be further improved by developments in hardware optimization, ethical AI methods, and training methodologies. However, there is still room for development in areas like energy consumption, interpretability, and bias prevention.These models will surely change industries and alter human-AI collaboration as they are integrated into real-world applications, from healthcare and education to entertainment and business. In summary, DeepSeek, ChatGPT, and Google Gemini are important turning points in the evolution of AI, each making a distinct contribution to the field. By understanding their distinct features and potential, we can better utilize their capabilities to address complex challenges and unlock new opportunities in the years to come.
\vspace{4mm}

\subsection*{\textbf{Data Availability:}} 
The datasets used and analyzed during the current study are available from the corresponding author on a reasonable request.
\vspace{2mm}

\subsection*{\textbf{Acknowledgments:}}
The authors thank the Research Ustad Team for supporting this research. 
\vspace{2mm}

\subsection*{\textbf{Grants:}}  
The authors received no funding for this study.
\vspace{2mm}

\subsection*{\textbf{Disclosures:}} 
The authors declare no conflicts of interest, financial or otherwise. 
\vspace{2mm}

\subsection*{\textbf{Author Contributions:}} 
All authors were involved in interpreting data, drafting the article, and revising it critically. All have approved the submitted and final versions.

\bibliographystyle{IEEEtran}

\bibliography{sample}

\end{document}